\documentclass[10pt,twocolumn,letterpaper]{article}

\usepackage{cvpr}
\usepackage{times}
\usepackage{epsfig}
\usepackage{graphicx}
\usepackage{amsmath}
\usepackage{amssymb}
\usepackage{subfig}
\usepackage{bm}
\usepackage{color}

\definecolor{darkblue}{rgb}{0,.1,.7}
\cvprfinalcopy
\begin{document}

\title{\vspace{-12mm}Learning an Interactive Segmentation System}

\author{Hannes Nickisch\\
Max Planck Institute for Biological Cybernetics\\
Spemannstr. 38, 72076 T\"ubingen, Germany\\
{\tt\small hn@tue.mpg.de}
\and
Pushmeet Kohli\\
Microsoft Research Cambridge\\
7 J J Thomson Avenue, Cambridge, CB3 0FB\\
{\tt\small pkohli@microsoft.com}
\and
Carsten Rother\\
Microsoft Research Cambridge\\
7 J J Thomson Avenue, Cambridge, CB3 0FB\\
{\tt\small carrot@microsoft.com}
}

\maketitle
\thispagestyle{empty}

\begin{abstract}
Many successful applications of computer vision to image or video manipulation are interactive by nature. However, parameters of such systems are often trained neglecting the user. Traditionally, interactive systems have been treated in the same manner as their fully automatic counterparts. Their performance is evaluated by computing the accuracy of their solutions under some fixed set of user interactions. This paper proposes a new evaluation and learning method which brings the user in the loop. It is based on the use of an active robot user - a simulated model of a human user. We show how this approach can be used to evaluate and learn parameters of state-of-the-art interactive segmentation systems. We also show how simulated user models can be integrated into the popular max-margin method for parameter learning and propose an algorithm to solve the resulting optimisation problem.
\end{abstract}

\section{Introduction} \label{sec:intro} Problems in computer vision
are known to be extremely hard, and very few fully automatic vision
systems exist which have been shown to be accurate and robust under
all sorts of challenging inputs. These conditions in the past had
made sure that most vision algorithms were confined to the
laboratory environment. The last decade, however, has seen computer
vision finally come out of the research lab and into the real world
consumer market. This great sea change has occurred primarily on the
back of the development of a number of interactive systems which
have allowed users to help the vision algorithm to achieve the
correct solution by giving hints. Some successful examples are
systems for image and video manipulation, and interactive 3D
reconstruction tasks. Image stitching and interactive image
segmentation are two of the most popular applications in this area.
Understandably, interest in interactive vision system has grown in
the last few years, which has led to a number of workshops and
special sessions in vision, graphics, and user-interface
conferences~\footnote{e.g. ICV07, and NIPS09}.

The performance of an interactive system strongly depends on a
number of factors, one of the most crucial being the user. This user
dependence makes interactive systems quite different from their
fully automatic counterparts, especially when it comes to learning
and evaluation. Surprisingly, there has been little
work in computer vision or machine learning devoted to {\em learning}
interactive systems. This paper tries to bridge this gap.

We choose interactive image segmentation to demonstrate the efficacy
of the ideas presented in the paper. However, the theory is general
and can be used in the context of any interactive system.
Interactive segmentation aims to separate a part of the image (an
object of interest) from the rest. It is treated as a classification
problem where each pixel can be assigned one of two labels: foreground (fg)
or background (bg). The interaction comes in the form of sets of pixels
marked by the user by help of brushes to belong either to fg or bg.
We will refer to each user interaction in
this scenario as a {\em brush stroke}.

This work addresses two questions: (1) How to evaluate any given
interactive segmentation system? and (2) How to learn the {\em best}
interactive segmentation system? Observe that the answer to the
first question gives us an answer to the second. One may imagine a
learning algorithm generating a number of possible segmentation
systems. This can be done, for instance, by changing parameter
values of the segmentation algorithm. We can then evaluate all such
systems, and output the best one.

We demonstrate the efficacy of our evaluation methods by learning
the parameters of the state-of-the-art system for interactive image
segmentation and its variants. We then go further, and show how the
max-margin method for learning parameters of fully automated
structured prediction models can be extended to do learning with the
user in the loop. To summarize, the contributions of this paper are:
(1) The study of the problems of evaluating and learning interactive
systems. (2) The proposal and use of a user model for evaluating and
learning interactive systems. (3) The first thorough comparison of
state-of-the-art segmentation algorithms under an explicit user
model. (4) A new algorithm for max-margin learning with user in the
loop.

\paragraph{Organization of the paper} In Section \ref{sec:eval}, we
discuss the problem of system evaluation. In Section
\ref{sec:segment}, we give details of our problem setting, and
explain the segmentation systems we use for our evaluation. Section
\ref{sec:linesearch} explains the na\"ive line-search method for
learning segmentation system parameters. In Section
\ref{sec:maxmargin}, we show how the max-margin framework for
structured prediction can be extended to handle
interactions, and show some basic results. The conclusions are given
in Section \ref{sec:conclusion}.

\section{Evaluating Interactive Systems} \label{sec:eval}
Performance evaluation is one of the most important problems in the
development of real world systems. There are two choices to be made:
(1) The data sets on which the system will be tested, and (2)
the quality or accuracy measure. Traditional
computer vision and machine learning systems are evaluated on
preselected training and test data sets. For instance, in automatic 
object recognition, one minimizes the number of misclassified pixels 
on datasets such as PASCAL VOC~\cite{pascal-voc}.

In an interactive system, these choices are much harder to make
because of the presence of an active user in the loop. Users behave
differently, prefer different interactions, may have different error
tolerances, and may also learn over time. The true objective
function of an interactive system -- although intuitive -- is hard
to express analytically: The user wants to achieve a satisfying
result easily and quickly. We will now discuss a number of possible
solutions, some of which, are well known in the literature.

\subsection{Static User Interactions} This is one of the most
commonly used methods in papers on interactive image
segmentation~\cite{RotherECCV04,singaraju09pbrush,duchenne08transductive}.
It uses a fixed set of user-made interactions (brush strokes)
associated with each image of the dataset. These strokes are mostly
chosen by the researchers themselves and are encoded using image
trimaps. These are pixel assignments with foreground, background,
and unknown labels (see Figure \ref{fig:robot_trimap}). The system
to be evaluated is given these trimaps as input and  their accuracy
is measured by computing the Hamming distance between the obtained
result and the ground truth. This scheme of evaluation does not
consider how users may change their interaction by observing the
current segmentation results. Evaluation and learning methods which
work with a fixed set of interactions will be referred to as {\em
static} in the rest of the paper.

Although the static evaluation method is easy to use, it suffers
from a number of problems: (1) The fixed interactions might be very
different from the ones made by actual users of the system. (2)
Different systems prefer different type of user hints (interaction
strokes) and thus a fixed set of hints might not be a good way of
comparing two competing segmentation systems. For instance, geodesic
distance based approaches
\cite{Bai:ICCV07,grady06randomWalker,singaraju09pbrush} prefer brush
strokes which are equidistant from the segmentation boundary as
opposed to graph cuts based
approaches~\cite{Boykov:ICCV2001,RKB:SIGGRAPH04}. (3) The evaluation
does not take into account how the accuracy of the results improves
with more user strokes. For instance, one system might only need a
single user interaction to reach the ground truth result, while the
other might need many interactions to get the same result. Still,
both systems will have equal performance under this scheme. These
problems of static evaluation make it a poor tool for judging the
relative performance of newly proposed segmentation system.

\subsection{User Studies} A user study involves the system being
given to a group of participants who are required to use it to solve
a set of tasks. The system which is easiest to use and yields the
correct segmentation in the least amount of time is considered the
best. Examples are \cite{li04lazysnapping} where a full user study
has been done, or \cite{Bai:ICCV07} where an advanced user
has done with each system the optimal job for a few images.

While overcoming most of the problems of a static
evaluation, we have introduced new ones: (1) User studies
are expensive and need a large number of participants to be of
statistical significance. (2) Participants need to be given enough
time to familiarize themselves with the system. For instance, an
average driver steering a Formula 1 car for the first time, might be
no faster than with a normal car. However, after gaining
experience with the car, one would expect the driver to be much
faster. (3) Each system has to be evaluated
independently by participants, which makes it infeasible to use this
scheme in a learning scenario where we are trying to find the
optimal parameters of the segmentation system among thousands or
millions of possible ones.

\subsection{Evaluation using Crowdsourcing} Crowdsourcing has attracted 
a lot of interest in the machine learning and computer vision 
communities. This is primarily due the success of a number of money
\cite{sorokin08mechTurk}, reputation \cite{ahn04ESP}, and community
\cite{russel08labelme} based incentive schemes for collecting
training data from users on the web. Crowdsourcing has the potential
to be an excellent platform for evaluating interactive vision
systems such as those for image segmentation. One could imagine
asking Mechanical Turk \cite{mech-turk} users to cut out different 
objects in images with different systems. The one
requiring the least number of interactions on average might be
considered the best. However, this approach too, suffers from
a number of problems such as fraud prevention. Furthermore, as in
user-studies, this cannot be used for learning in light of thousands
or even millions of systems.

\subsection{Evaluation with an Active User Model} In this paper we
propose a new evaluation methodology which overcomes most of the
problems described above. Instead of using a fixed set of
interactions, or an army of human participants, our method only
needs a model of user interactions. This model is a simple algorithm
which -- given the current segmentation, and the ground truth --
outputs the next user interaction. This user model can be coded up
using simple rules, such as  ``give a hint in the middle of the
largest wrongly labelled region in the current solution'', or
alternatively, can be learnt directly from the interaction logs
obtained from interactive systems deployed in the market. There are
many similarities between the problem of learning a user model and
the learning of an agent policy in reinforcement learning. Thus, one
may exploit reinforcement learning methods for this task. Pros and
cons of evaluation schemes are summarized in Table
\ref{tab:ialearn_methods}.

\begin{table}[t!]
\centering
\resizebox{\columnwidth}{!}{
\begin{tabular}{|c||c|c|c|c|c|c|c|}
\hline
{\bf Method} & user  & user can & inter- & effort  & parameter  & time & price\tabularnewline
 & in loop & learn & action & model & learning &  & \tabularnewline
\hline
\hline
User model & yes & yes & yes & yes & this paper & fast & low\tabularnewline
\hline
Crowd sourcing & yes & yes & yes & yes & conceivable & slow & a bit\tabularnewline
\hline
User study & yes & yes & yes & yes & infeasible & slow & very high\tabularnewline
\hline
Static learning & no & no & no & no & used so far & fast & very low\tabularnewline
\hline
\end{tabular}
} \caption{Comparison of methods for interactive learning.}
\label{tab:ialearn_methods}
\end{table}

\section{Image Segmentation: Problem Setting} \label{sec:segment}

\begin{figure}[h]
\subfloat[Images $\mathbf{x}^k$]{\begin{centering}
 \includegraphics[angle=90,width=0.26\columnwidth]{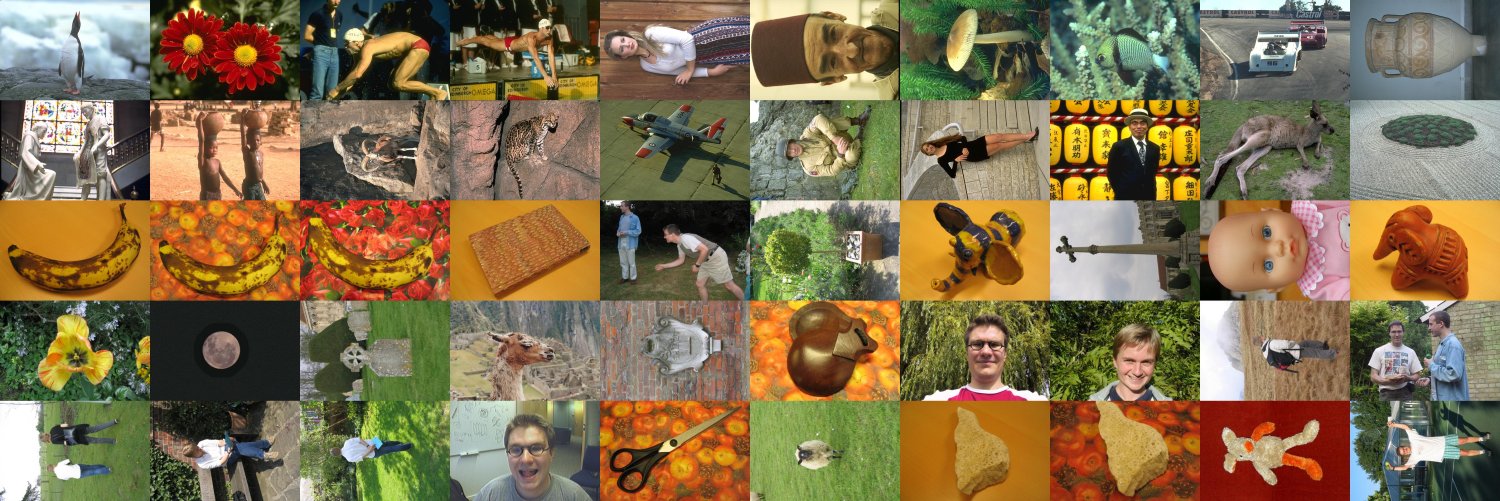}
 \par\end{centering}
 }\hfill{}
\subfloat[User trimaps $\mathbf{u}^k$]{\begin{centering}
 \includegraphics[angle=90,width=0.26\columnwidth]{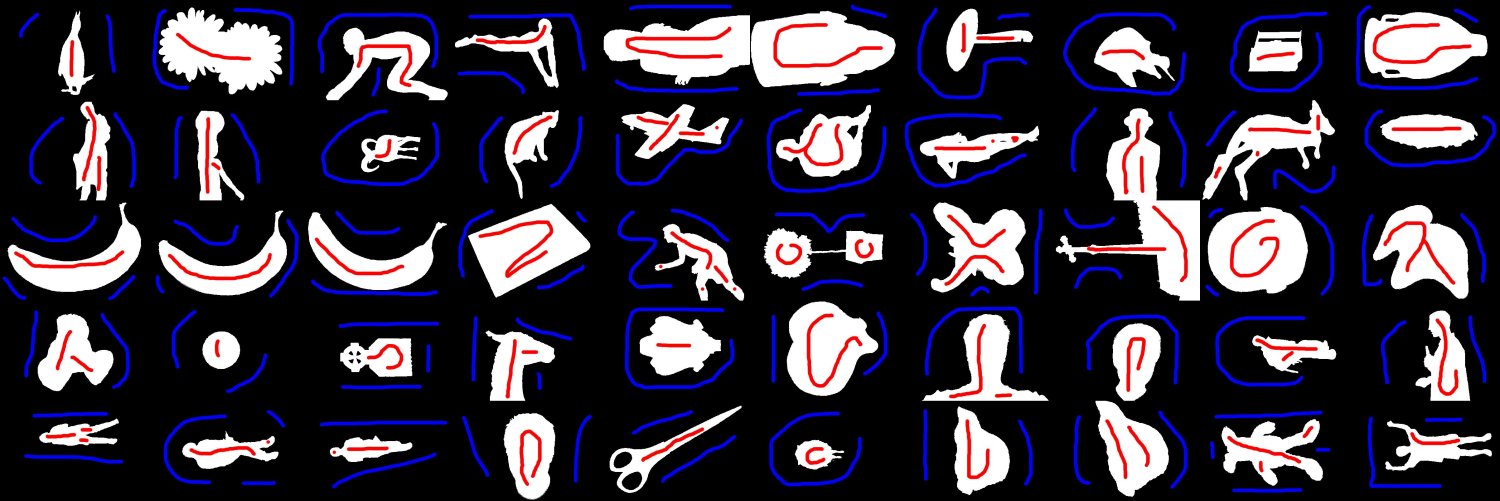}
 \par\end{centering}
 }\hfill{}
\subfloat[Tight trimaps $\mathbf{u}^k$]{\begin{centering}
 \includegraphics[angle=90,width=0.26\columnwidth]{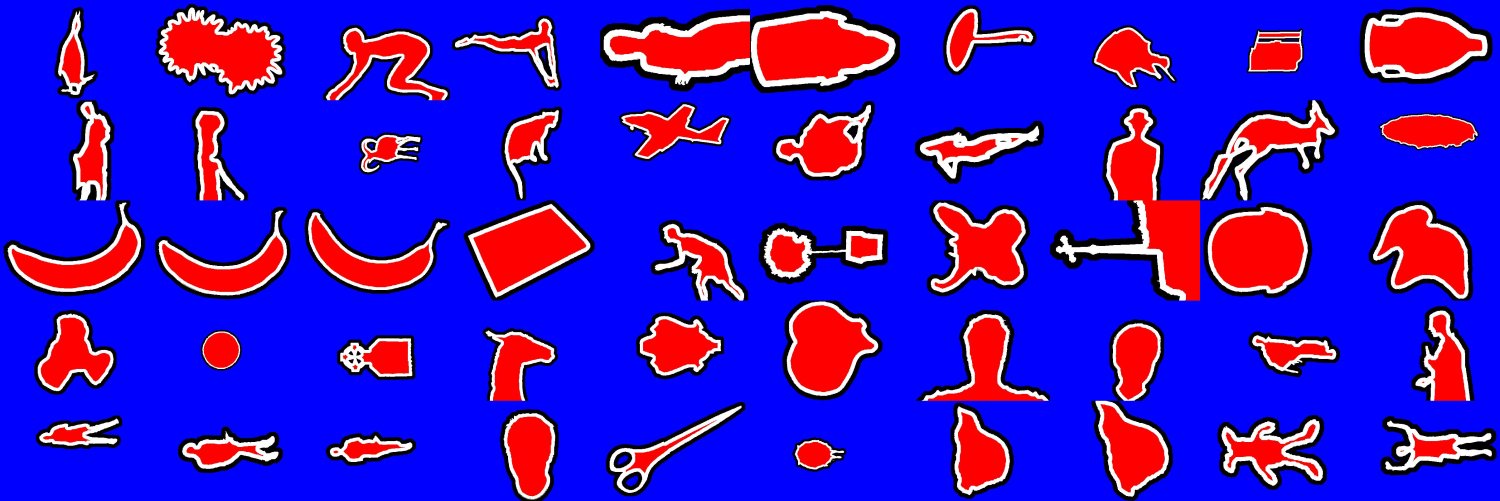}
 \par\end{centering}
 }

\caption{\label{fig:dataset}As detailed in the paper, we took the 50 GrabCut images \cite{GrabCut} with given ground truth segmentations (coded as black/white).
We considered two kinds of user inputs (codes as red/blue) : User defined strokes and tight trimaps generated by eroding the groundtruth segmentation. The user strokes where drawn by only looking at the ground truth segmentation $\mathbf{y}^k$ and ignoring the image $\mathbf{x}^k$.
}
\end{figure}

\subsection{The Database} \label{sec:database} 
\begin{figure*}[!th]
\subfloat[Input image]{\begin{centering}
\includegraphics[width=0.4\columnwidth]{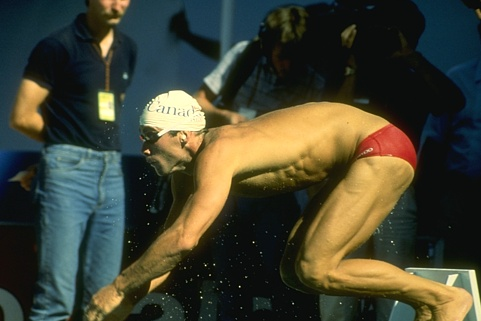}
\par\end{centering}

}\hfill{}\subfloat[\label{fig:robot_trimap}Tight trimap]{\begin{centering}
\includegraphics[width=0.4\columnwidth]{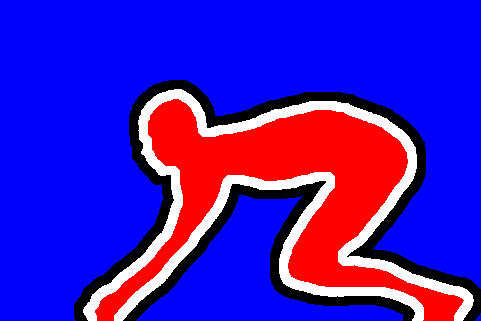}
\par\end{centering}

}\hfill{}\subfloat[\label{fig:robot_stroke}Robot user, $er\negmedspace=\negmedspace1.4\%$]{\begin{centering}
\includegraphics[width=0.4\columnwidth]{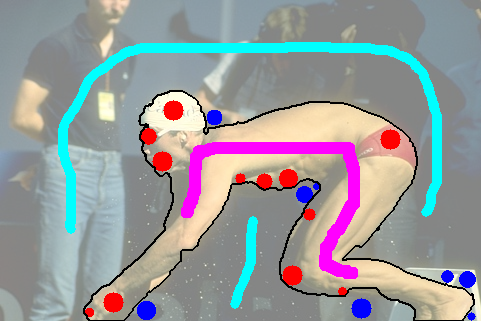}
\par\end{centering}

}\hfill{}\subfloat[\label{fig:robot_compare}Different robot users]{\begin{centering}
\includegraphics[width=0.75\columnwidth]{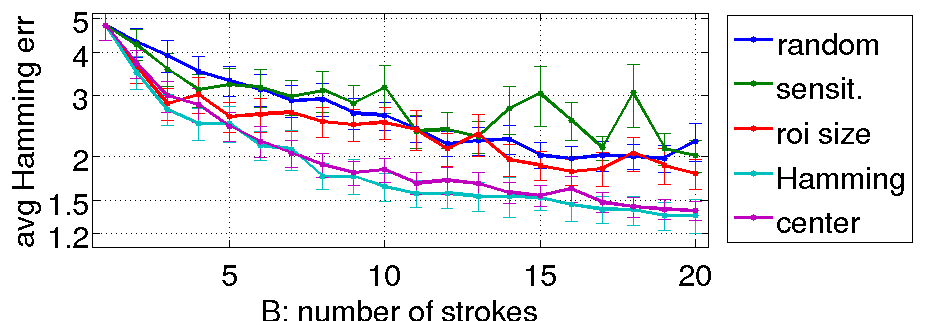}
\par\end{centering}

}
\caption{\label{fig:robot}An image from the database (a), tight trimap (b), robot user (red/blue) started from user scribbles (magenta/cyan) with segmentation (black) after $B\negmedspace=\negmedspace20$ strokes (c) and segmentation performance comparison of different robot users (d).
}
\end{figure*}

We use the publicly
available GrabCut database of 50 images, in which ground truth
segmentations are known \cite{GrabCut}.
In order to perform large scale testing and comparison, we
down-scaled all images to have a maximum size of $241 \times 161$,
while keeping the original aspect ratio\footnote{We confirmed by
visual inspection that the quality of segmentation results is not
affected by this down-scaling operation.}. For each image, we
created two different {\bf static} user inputs: (1) A ``static
trimap'' computed by dilating and eroding the ground truth
segmentation by $7$ pixels\footnote{This kind of input is used by
most systems for both comparison to competitors and learning of
parameters, e.g. \cite{RotherECCV04,singaraju09pbrush}.}. (2) A
``static brush'' consisting of a few user made brush strokes which
very roughly indicate foreground and background. We used on average
about $4$ strokes per image. (The magenta and cyan strokes in Fig.
\ref{fig:robot_stroke} give an example). All this data is visualized
in Figure \ref{fig:dataset}. 
Note, in Sec. \ref{sec:robuser} we will describe a
third ``dynamic trimap'' called the {\em robot user} where we
simulate the user.

\subsection{The Segmentation Systems} \label{sec:segsys} We now
describe 4 different interactive segmentation systems we use
in the paper. These are: ``GrabCutSimple(GCS)'', ``GrabCut(GC)'',
``GrabCutAdvanced(GCA)'', ``GeodesicDistance'' (GEO).

GEO is a very simple system. We first learn Gaussian
Mixture Model (GMM) based color models for fg/bg from user made brush strokes. We then
simply compute the shortest path in the likelihood ratio image as
described in \cite{Bai:ICCV07} to get a segmentation.

The other three systems all built on graph cut. They all work by
minimizing the energy function:
\begin{equation}
E(\mathbf{y})=\sum_{p\in \mathcal{V}} E_p(y_p) + \sum_{(p,q)\in \mathcal{E}} E_{pq}(y_p,y_q)
\label{eq:E}
\end{equation}
Here $(\mathcal{V},\mathcal{E})$ is an undirected graph whose nodes correspond
to pixels, $y_p\in\{0,1\}$ is the segmentation label of image pixel
$p$ with color $x_p$, where 0 and 1 correspond to the background and
the foreground respectively. We define $(\mathcal{V},\mathcal{E})$ to be an
8-connected 2D grid graph.

The unary terms are computed as follows. A probabilistic model is
learnt for the colors of background
($y_p\negmedspace=\negmedspace0$) and foreground
($y_p\negmedspace=\negmedspace1$) using two different GMMs
$\textrm{Pr}(x|0)$ and $\textrm{Pr}(x|1)$. $E_p(y_p)$
is then computed as $-\log(\textrm{Pr}(x_p|y_p))$ where $x_p$
contains the three color channels of pixel $p$. An important concept
of GrabCut \cite{RKB:SIGGRAPH04} is to update the color models based
on the whole segmentation. In practice we use a few iterations.

The pairwise term incorporates both an Ising prior and a
contrast-dependent component and is computed as
\begin{equation*}
E_{pq}(y_p,y_q) = \frac{\left|y_q - y_p\right|}{\textrm{dist}\left(p,q\right)} \left( w_i + w_c \exp \left[ -\beta { \left\| {x_p-x_q} \right\| }^2 \right] \right)
\end{equation*}
where $w_i$ and $w_c$ are weights for the Ising and
contrast-dependent pairwise terms respectively, and  \mbox{$\beta =
0.5 \cdot w_\beta / {\left\langle{ \left\| {x_p-x_q} \right\|
}^2\right\rangle}$} is a parameter, where $\left\langle \cdot
\right\rangle$ denotes expectation over an image sample
\cite{RKB:SIGGRAPH04}. We can scale $\beta$ with the parameter
$w_\beta$.

To summarize, the models have two linear free parameters: $w_i, w_c$
and a single non-linear one: $w_\beta$. The system GC minimizes the
energy defined above, and is pretty close to the original GrabCut
system~\cite{RKB:SIGGRAPH04}. GrabCutSimple(GCS) is a simplified
version, where color models (and unary terms) are fixed up front;
they are learnt from the initial user brush strokes (see Sec.
\ref{sec:segsys}) only. GCS will be used in max-margin learning and
to check the active user model, but it is not considered as a
practical system.

Finally, ``GrabCutAdvanced(GCA)'' is an advanced GrabCut system
performing considerably better than GC. Inspired by recent work
\cite{liu09paintselect}, foreground regions are 4-connected to a
user made brush stroke to avoid deserted foreground islands.
Unfortunately, such a notion of connectivity leads to an NP-hard
problem and various solutions have been
suggested~\cite{vicente08connectivity,nowozin09global}. However, all
these are either very slow and operate on super-pixels
\cite{nowozin09global} or have a very different interaction
mechanism~\cite{vicente08connectivity}. We simply remove deserted
foreground islands in a postprocessing step.

\subsection{The Robot User} \label{sec:robuser} We now describe
the different active user models tested and deployed by us. Given
the ground truth segmentation $\mathbf{y}^k$ and the current 
segmentation solution $\mathbf{y}$, the active user model is a policy 
$s:(\mathbf{x}^k,\mathbf{y}^k,\mathbf{u}^{k,t},\mathbf{y})\mapsto\mathbf{u}^{k,t+1}$ 
which specifies which brush stroke to place next. Here, $\mathbf{u}^{k,t}$
denotes the user interaction history of image $\mathbf{x}^{k}$ up to 
time $t$. We have investigated various options for this policy:
(1) Brush strokes at random image positions. (2) Brush strokes in the
middle of the wrongly labelled region (center). For the second
strategy, we find the largest connected region of the binary mask,
which is  given by the absolute difference between the current
segmentation and ground truth. We then mark a circular brush stroke
at the pixel which is inside this region and furthest away from the
boundary. This is motivated by the observation that users tend to
find it hard to mark pixels at the boundary of an object because
they have to be very precise.

We also tested user models which took the segmentation algorithm
explicitly into account. This is analogous to users who have learnt
how the segmentation algorithm works and thus interact with it
accordingly. We consider the user model which marks a circular brush
stroke at the pixel (1) with the lowest min marginal (sensit). (2)
which results in the largest change in labeling (roi size). (3)
which decreases the Hamming error by the biggest amount (Hamming).
We consider each pixel as the circle center and choose the one where
the Hamming error decreases most (Hamming). This is very expensive,
but in some respects is the best solution\footnote{Note, one could
do even better by looking at two or more brushes after each other
and then selecting the optimal one. However, the solution grows
exponentially with the number look-ahead steps.}. ``Hamming'' acts
as a very ``advanced user'', who knows exactly which interactions
(brush strokes) will reduce the error by the largest amount. It is
quite questionable that a user is actually able to find the optimal
position, and a user study might be needed. On the other hand, the
``centre'' user model behaves as a ``novice user''.

Fig. \ref{fig:robot_stroke} shows the result of a robot user
interaction, where cyan and magenta are the initial fixed brush
strokes (called ``static brush trimap''), and the red and blue dots
are the robot user interactions. The robot sets brushes of a maximum
fixed size (here 4 pixel radius). Apart from the true object
boundary, the maximum brushes size is used. At the boundary, the
brush size is scaled down, in order to avoid that the brush
straddles the boundary.

Fig. \ref{fig:robot_compare} shows the performance of the 5
different user models (robot users) over a range of $20$ brushes.
Here we used the GCS system, since it is computationally infeasible
to apply the (sensit; roi; Hamming) user models on other interaction
systems. GCS can be used because it allows efficient computation of
solutions. It does this by recycling computation when doing the
optimization \cite{kohli05dynamicGC}. In the other systems, this is not
possible, since unaries change with every brush stroke, and hence we
have to treat the system as a black box.

As expected, the random user performs badly. Interestingly the robot
users minimizing the energy (roi, sensit) also perform
badly. Both ``Hamming'' and ``centre'' are considerably better than
the rest. It is interesting to note that ``centre'' is actually
only marginally worse than ``Hamming''. It has to be said that for
other systems, e.g. GEO this conclusion might not hold, since e.g.
GEO is much sensitive to the location of the brush stroke than a
system based on graph cut, as \cite{singaraju09pbrush} has shown. To
summarize, ``centre'' is the robot user strategy which simulates a
``novice user'' and is computational feasible, since it does not
look at the underlying system at all. Also, ``centre'' performed for
GCS nearly the same as the optimal strategy ``Hamming''. Hence, for
the rest of the paper we always stick to the user (centre) which we
call from here onwards our {\it robot user}.

\subsection{The Error Measure} \label{sec:error} For a static trimap
input there are many different ways for obtaining an error rate, see
e.g. \cite{RotherECCV04,Kohli08}. In a static setting, most papers
use the number of misclassified pixels (Hamming distance) between
the ground truth segmentation and the current result. We call this
measure ``$er_b$'', i.e. Hamming error for brush $b$. One could do
variations, e.g. \cite{Kohli08} weight distances to the boundary
differently, but we have not investigated this here. Fig.
\ref{fig:robot_compare} shows how the Hamming error behaves with
each interaction.

For learning and evaluation we need an error metric giving
us a single score for the whole interaction. One choice is the
``weighted'' Hamming error averaged over a fixed number of brush
strokes B. In particular we choose the error ``$Er$'' as: $Er = [
\sum_b f(er_b) ] / B$ where $f(er_b) = er_b$. Note, to ensure a fair
comparison between systems, B must be the same number for all
systems. Another choice for the quality metric which matches more
closely with what the user wants is described as follows. We use a
sigmoid function $f:\mathbb{R}_{+}\rightarrow[0,c]$ of the form:
\begin{equation} f(e)=\begin{cases}
0 & er_b\le1.5\\
c-\frac{c}{(er_b-0.5)^{2}} & er_b>1.5\end{cases},\quad
c=5\label{eq:transfer_function}\end{equation}

Observe that $f$ encodes two facts: all errors below $1.5$ are
considered negligible and large errors do never weigh more than $c$.
The first reasons of this settings is that visual inspection showed
that for most images, an error below $1.5\%$ corresponds to a
visually pleasing result. Of course this is highly subjective, e.g. a
missing limb from the segmentation of a cow might be an error of
$0.5\%$ but is visually unpleasing, or an incorrectly segmented
low-contrast area has an error of $2\%$ but is visually not
disturbing. The second reason for having a
maximum weight of $c$ is that users do not discriminate between two
systems giving large errors. Thus errors of 50\% and 55\% are equally penalized.

Due to runtime limitations for parameter learning, we do want to run
the robot user for not too many brushes (e.g. maximum of $20$
brushes).  Thus we start by giving an initial set of brush strokes
which are used to learn the colour models. At the same time, we want
that most images reach an error level of about $1.5\%$. When we
start with a static brush trimap we get for $68\%$ of images an
error rate smaller than $1.5\%$ and for $98\%$ smaller than $2.5\%$,
with the GCA system. We also confirmed that the inital static brush
trimap does not affect the learning considerably\footnote{We started
the learning from no initial brushes and let it run for 60 brush
strokes. The learned parameters were similar as with starting from
20 brushes}.

\section{Interactive Learning by line-search} \label{sec:linesearch}
Systems with few parameters can be trained by simple line (or grid)
search. Our systems, GC and GCA, have only 3 free parameters: $w_c,
w_i, w_\beta$. Line search is done by fixing all but one free
parameter $w_\phi$ and simulating the user interaction process for
$30$ different discrete values $w_{\phi,i}$ of the free parameter
$w_\phi$ over a predefined range. The optimal value $w_\phi^*$ from
the discrete set is chosen to minimize the leave-one-out (LOO)
estimator of the test error\footnote{This is number-of-data-point-fold cross
validation.}. Not only do we prevent
overfitting but we can efficiently compute the Jackknife estimator
of the variance \cite[ch. 8.5.1]{wasserman04allOfStat}, too. This
procedure is done for all parameters in sequence with a
sensible starting point for all parameters. We do one sweep only.
One important thing to notice is that our dataset was big enough
(and our parameter set small enough) as to not suffer from
over-fitting. We see this by observing that training and test error
rates are virtually the same for all experiments. In addition to the
optimal value we obtain the variance for setting this parameter. In
rough words, this variance tells us, how important it is to have
this particular value. For instance, a high variance means that
parameters different from the selected one, would also perform well.
Note, since our error function (Eq. \ref{eq:transfer_function}) is
defined  for both, trimaps which are static and dynamic, the above
procedure can be performed for all three different types of trimaps:
``static trimap'', ``static brush'', ``dynamic brush''.

\begin{figure*}[!ht]
\subfloat[\label{fig:grid_search_GCA_wc}GCA, $w_c$ training]{\begin{centering}
\includegraphics[width=0.49\columnwidth]{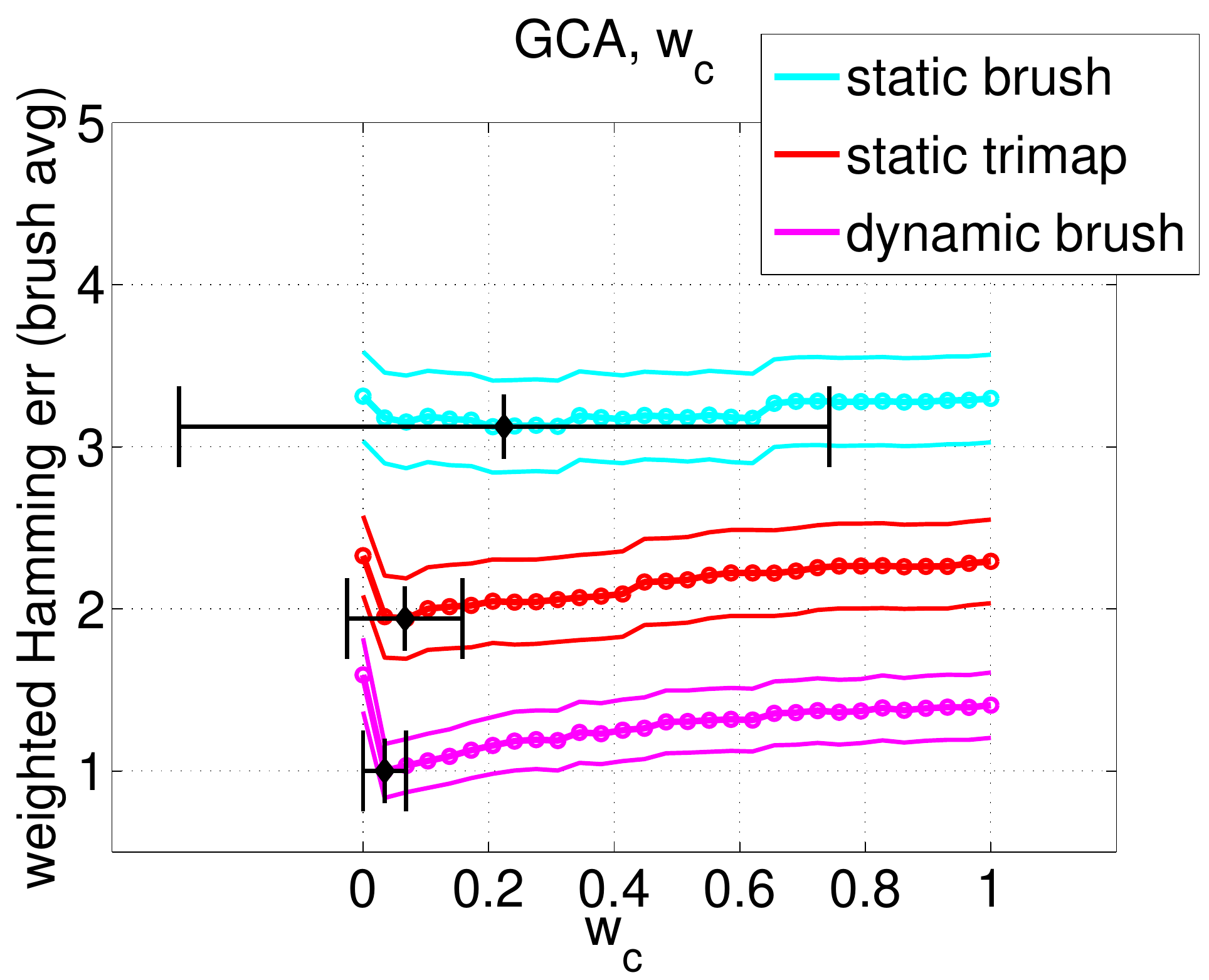}
\par\end{centering}

}\hfill{}\subfloat[GCA, $w_\beta$ training]{\begin{centering}
\includegraphics[width=0.49\columnwidth]{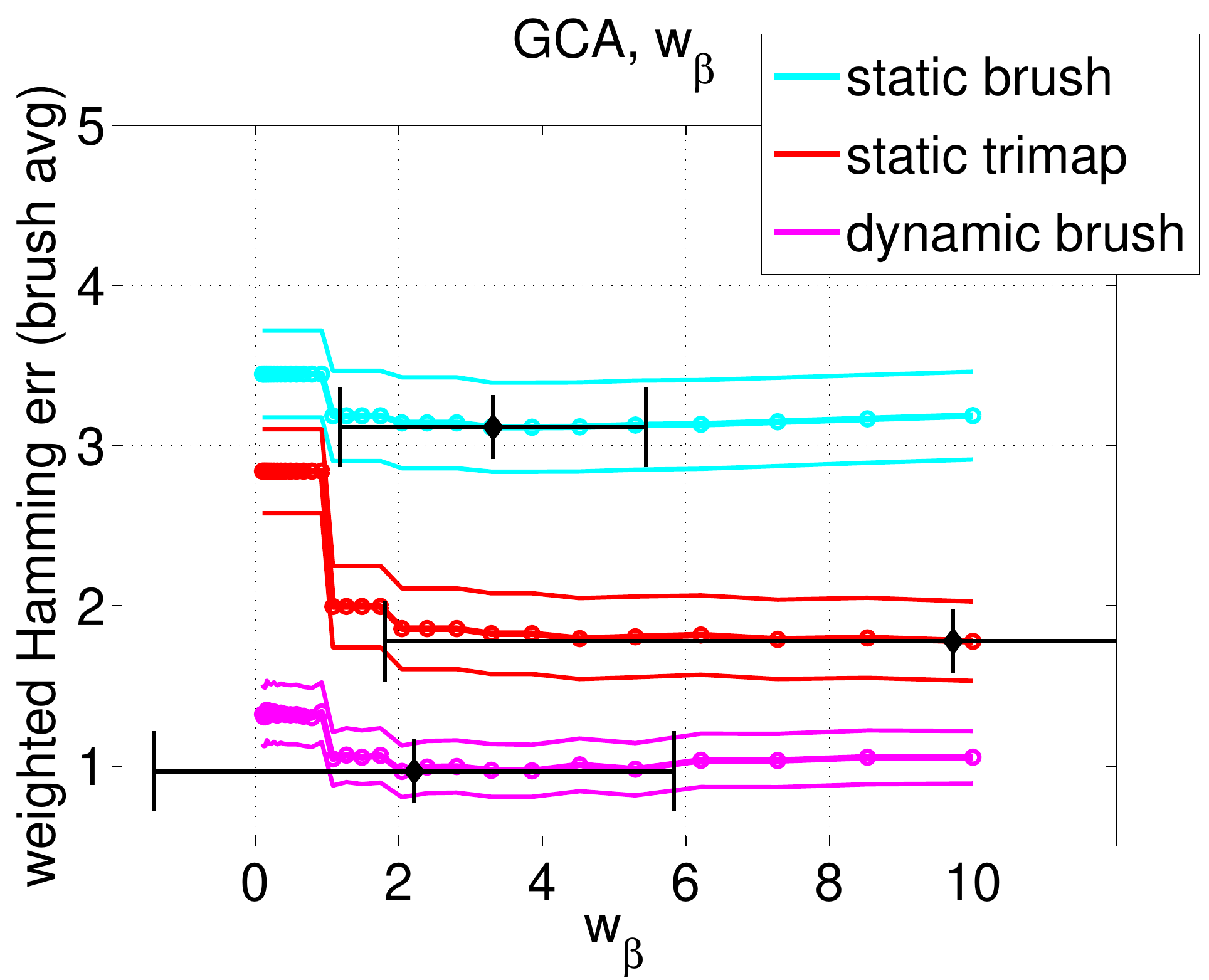}
\par\end{centering}

}\hfill{}\subfloat[GC, $w_c$ training]{\begin{centering}
\includegraphics[width=0.49\columnwidth]{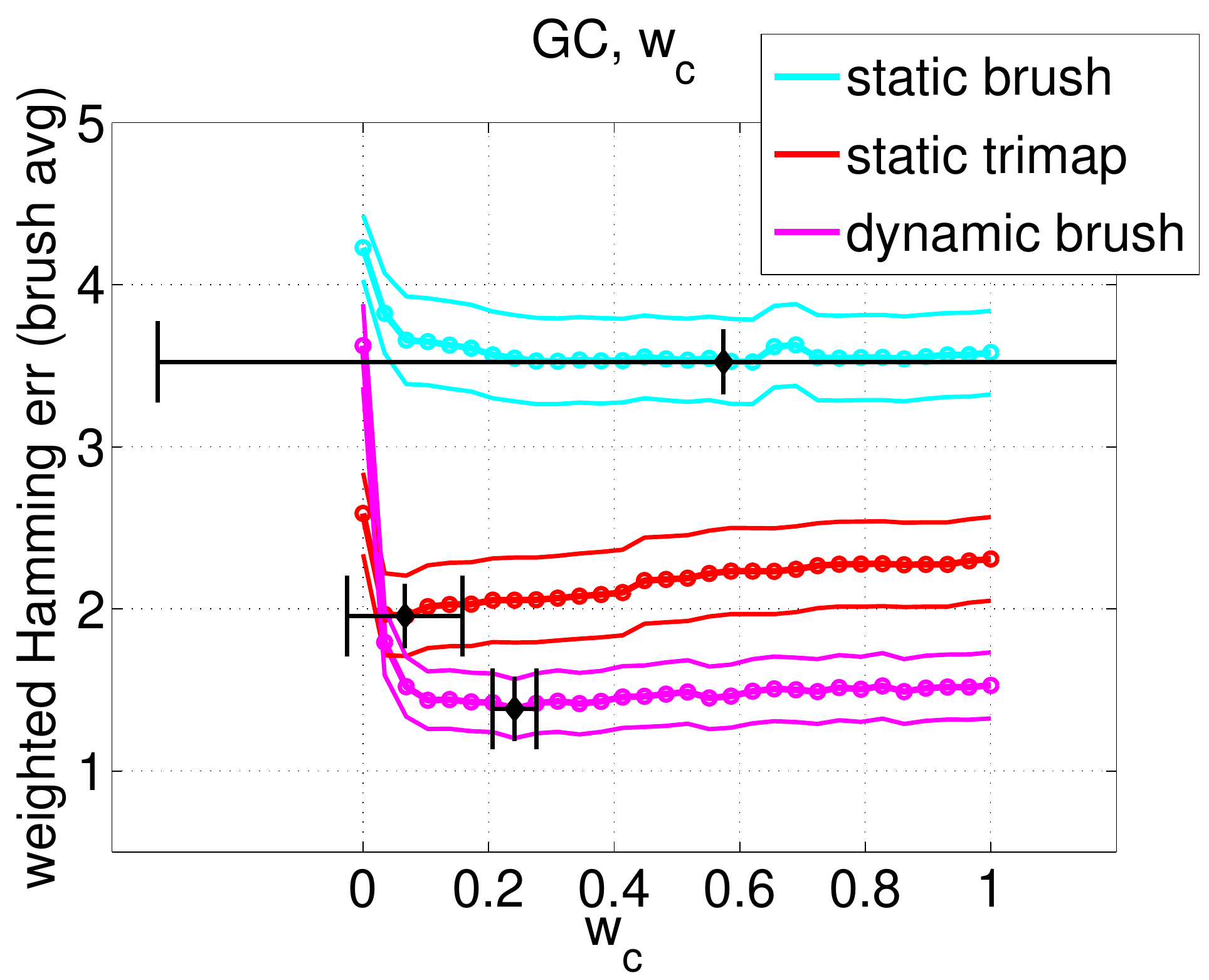}
\par\end{centering}

}\hfill{}\subfloat[GCA, $(w_c,w_\beta)$ test]{\begin{centering}
\includegraphics[width=0.49\columnwidth]{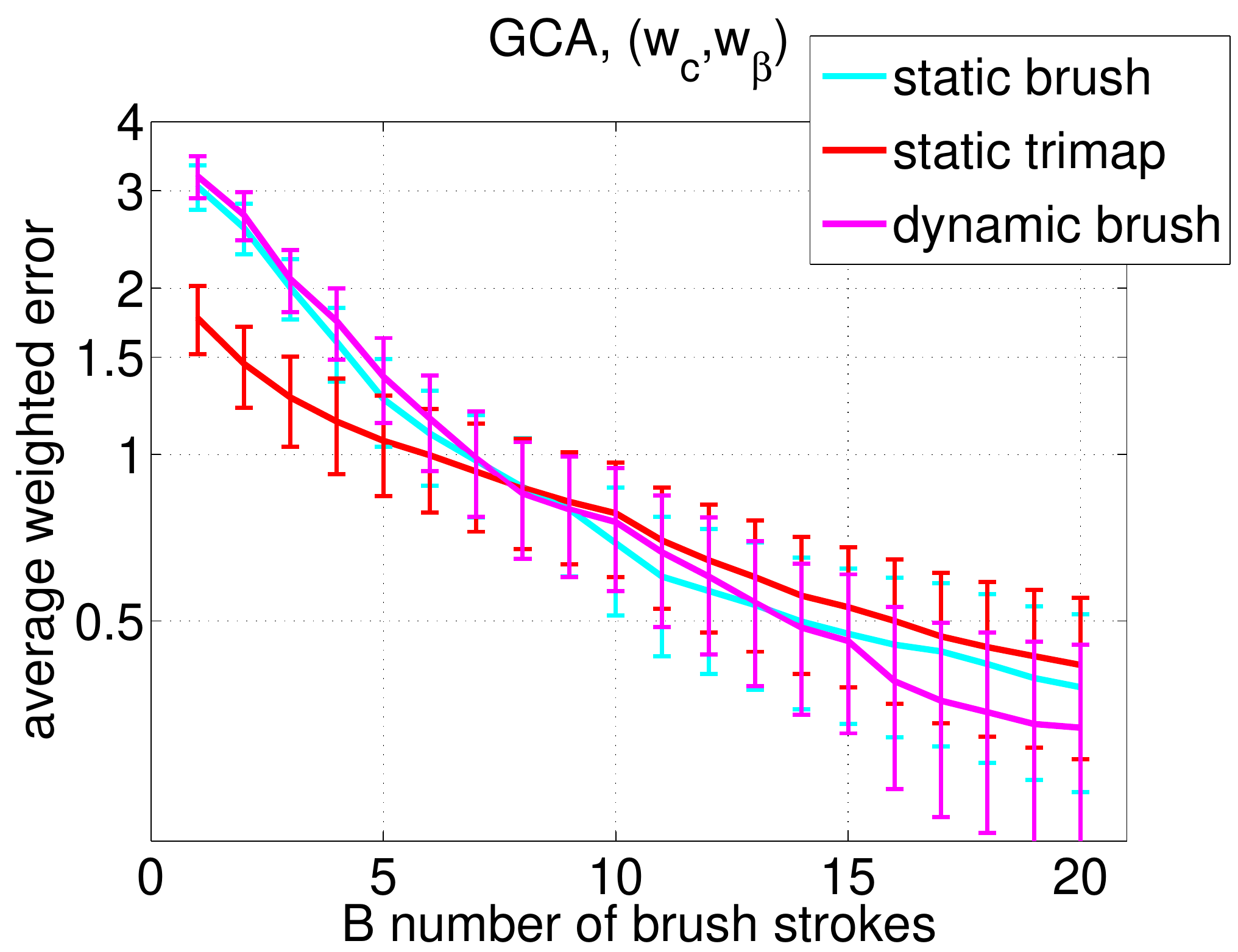}
\par\end{centering}

}
\caption{\label{fig:grid_search}
Line search. We compare $3$ different training procedures for interactive segmentation: Static learning from a fixed set of user brushes, static learning from a tight trimap and dynamic learning with a robot user starting from a fixed set of user brushes. Train (a-c): Reweighted Hamming errors ($\pm$ stdev.) for two segmentation systems (GC/GCA) as a function of two line-search parameters ($w_c$,$w_\beta$). The optimal parameter is shown along with its Jackknife variance estimate (black horizontal bar). Test (d): Segmentation performance using the optimal parameters $(w_c^*,w_\beta^*)$ after iterated line search optimisation. Note that the dynamically learnt paramters develop their strength in the course of interaction.
}
\end{figure*}

Table \ref{tab:LineSearchGCA} summaries all the results, and
Fig. \ref{fig:grid_search} illustrates some results during training
and test (caption explains details of the plots). One can observe
that the three different trimaps suggest different optimal
parameters for each system, and are differently certain about them.
This leads to  {\it key contribution} of this study:  A system which
is interactive in test time has also to be trained in an interactive
way. We see from the test plots that indeed, a system trained with
``dynamic trimap'' does better than trained with either ``static
brush'' or ``static trimap''.

Let us look closer at some learnt settings. For system GCA and
parameter $w_c$ (see Table \ref{tab:LineSearchGCA} (first row), and
Fig. \ref{fig:grid_search_GCA_wc} we observe that the optimal value
in a dynamic setting is lower ($0.03$) than in any of the static
settings. This is surprising since one would have guessed that the
true value of $w_c$ lies somewhere in between a loose and very tight
trimap. Interestingly in \cite{singaraju09pbrush}, the authors had
learned a parameter by averaging the performance from two static
trimaps. From the above study, one might have concluded the static
``tight trimap'' might give good insights about the choice of
parameters. However, when we now consider the training of the
parameter $w_\beta$ in the GCA system, we see that such a conclusion
would be wrong, since the ``tight trimap'' reaches a very different
minimum ($9.73$) than the dynamic trimap ($2.21$).\footnote{Note,
the fact that the uncertainty of the ``tight trimap'' learning is
high, gives an indication that this value can not be trusted very
much.} To summarize, conclusions about the optimal parameter setting
of an interactive system should be drawn by a large set of
interaction and cannot be made by looking solely at a few (here two)
static trimaps.

\begin{table}[ht!]
\centering
\resizebox{.9\columnwidth}{!}{
\begin{tabular}{c|ccc}
Trimap & $w_c$ & $w_i$ & $w_\beta$ \\
\hline
dynamic brush & 0.03$\pm$ 0.03 & 4.31$\pm$ 0.17 & 2.21$\pm$ 3.62 \\
static trimap & 0.07$\pm$ 0.09 & 4.39$\pm$ 4.40 & 9.73$\pm$ 7.92 \\
static brush & 0.22$\pm$ 0.52 & 0.47$\pm$ 8.19 & 3.31$\pm$ 2.13 \\
\hline
\end{tabular}
}
\caption{\label{tab:LineSearchGCA} System GCA. Optimal values $\pm$ stdev.}
\end{table}

For the sake of completeness, we have the same numbers for the GC
system in Table \ref{tab:LineSearchGC}. We see the same conclusions as above. One
interesting thing to notice here is that the pairwise terms (esp.
$w_c$) are chosen higher than in GCA. This is expected, since without
post-processing a lot of isolated islands may be present which are
far away from the true boundary. So post-processing automatically
removes these islands. The effect is that in GCA the pairwise terms
can now concentrate on modeling the smoothness on the boundary
correctly. However, in GC the pairwise terms have to additionally
make sure that the isolated regions are removed (by choosing a
higher value for the pairwise terms) in order to compensate for the
missing post-processing step.

\begin{table}[ht!]
\centering
\resizebox{.9\columnwidth}{!}{
\begin{tabular}{c|ccc}
Trimap & $w_c$ & $w_i$ & $w_\beta$ \\
\hline
dynamic brush & 0.24$\pm$ 0.03 & 4.72$\pm$ 1.16 & 1.70$\pm$ 1.11 \\
static trimap & 0.07$\pm$ 0.09 & 4.39$\pm$ 4.40 & 4.85$\pm$ 6.29 \\
static brush & 0.57$\pm$ 0.90 & 5.00$\pm$ 0.17 & 1.10$\pm$ 0.96 \\
\hline
\end{tabular}
}
\caption{\label{tab:LineSearchGC} System GC. Optimal values $\pm$ stdev.}
\end{table}

It is interesting to note that for the error metric $f(er_b)=er_b$, we get
slightly different values, see Table \ref{tab:LineSearchGCAcomplete}. For instance,
we see that $w_c = 0.07 \pm 0.07$ for GCA with our active user. This
is not too surprising, since it says that larger errors are more
important (this is what $f(er_b)=er_b$ does). Hence, it is better to
choose a larger value of $w_c$.

In Figure \ref{fig:grid_search}d of the paper we plot the actual segmentation error $f(er_b)$ and not the error measure $\sum_{b=1}^B f(er_b)$ for $f(er_b)=sigmoid(er_b)$. In Table \ref{tab:LineSearchGCAcomplete}, we have collected all final error measure values. It is visible from the table that the dynamically adjusted parameters only perform better in terms of the instantaneous error but not in terms of the cumulative error measure.

\begin{table}[ht!]
\centering
\resizebox{.95\columnwidth}{!}{
\begin{tabular}{|c||c|c|c|}
\hline 
 & static brush & static trimap & dynamic brush\tabularnewline
\hline
\hline 
$f(er_{b})=\text{sigmoid}(er_{b})$ [shown] & $0.379\pm0.134$ & $0.416\pm0.135$ & $0.321\pm0.132$\tabularnewline
\hline 
$f(er_{b})=\text{sigmoid}(er_{b}),\quad\sum_{b}f(er_{b})$ & $0.984\pm0.165$ & $0.820\pm0.173$ & $1.007\pm0.163$\tabularnewline
\hline 
$f(er_{b})=er_{b}$ & $1.191\pm0.072$ & $1.232\pm0.071$ & $1.160\pm0.074$\tabularnewline
\hline 
$f(er_{b})=er_{b},\quad\sum_{b}f(er_{b})$ & $1.564\pm0.103$ & $1.422\pm0.084$ & $1.610\pm0.115$\tabularnewline
\hline
\end{tabular}
}
\caption{\label{tab:LineSearchGCAcomplete} System GCA. Optimal values $\pm$ stdev.}
\end{table}

\begin{figure*}[h]
 \subfloat[GCA, contrast weight $w_c$]{\begin{centering}
 \includegraphics[width=0.6\columnwidth]{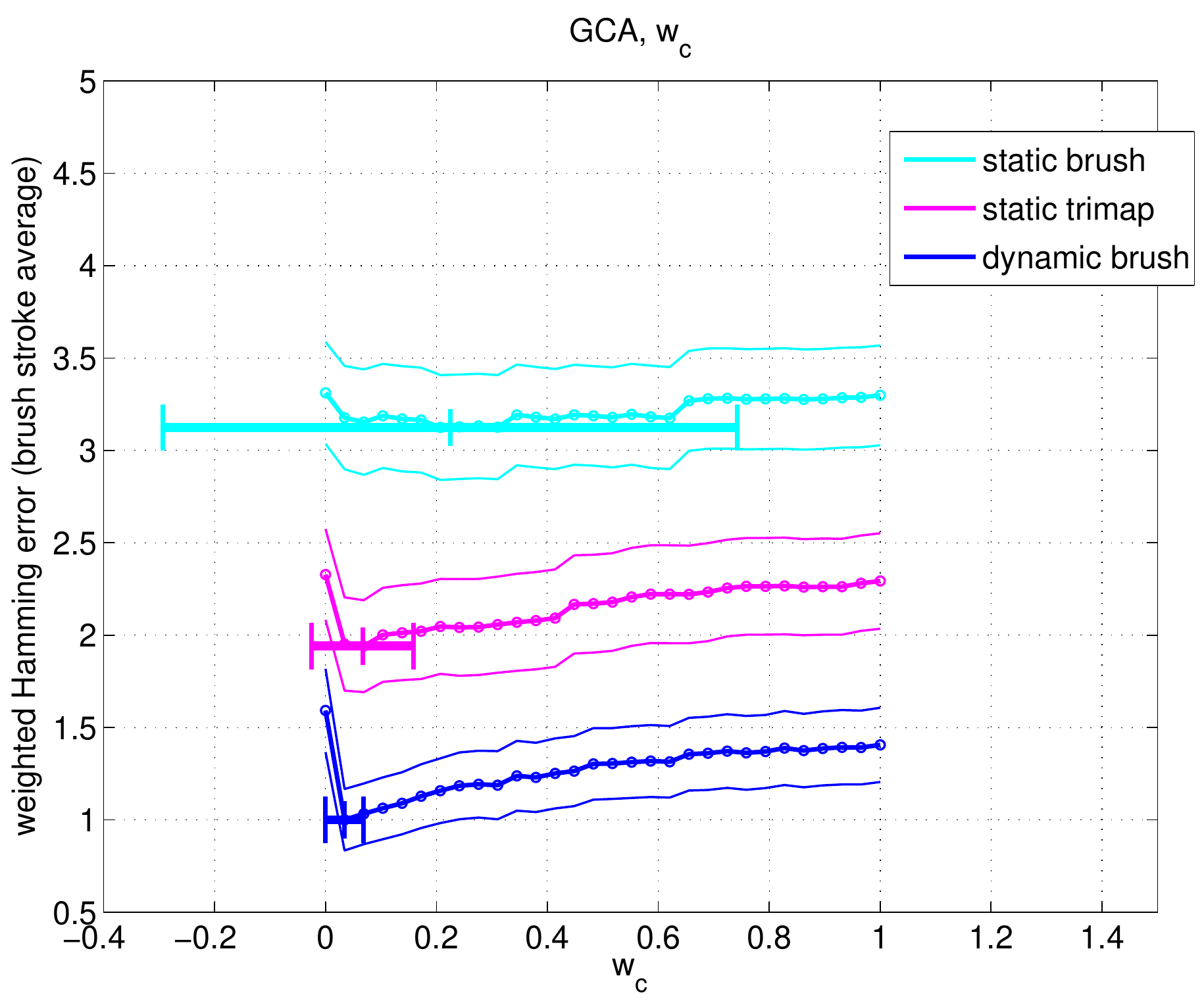}
 \par\end{centering}

 }\hfill{}\subfloat[GCA, Ising weight $w_i$]{\begin{centering}
 \includegraphics[width=0.6\columnwidth]{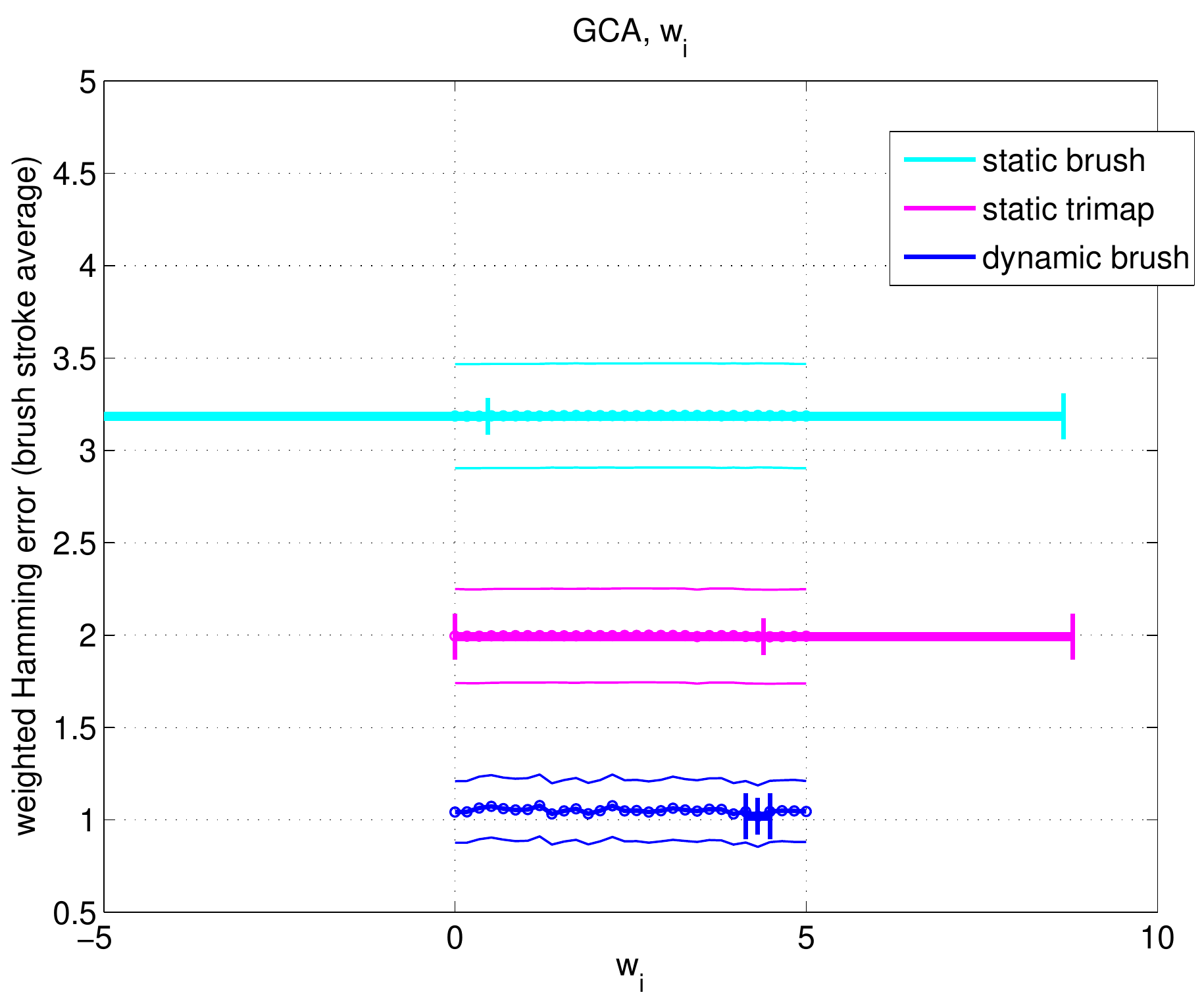}
 \par\end{centering}

 }\hfill{}\subfloat[GCA, $\beta$-scale $w_\beta$]{\begin{centering}
 \includegraphics[width=0.6\columnwidth]{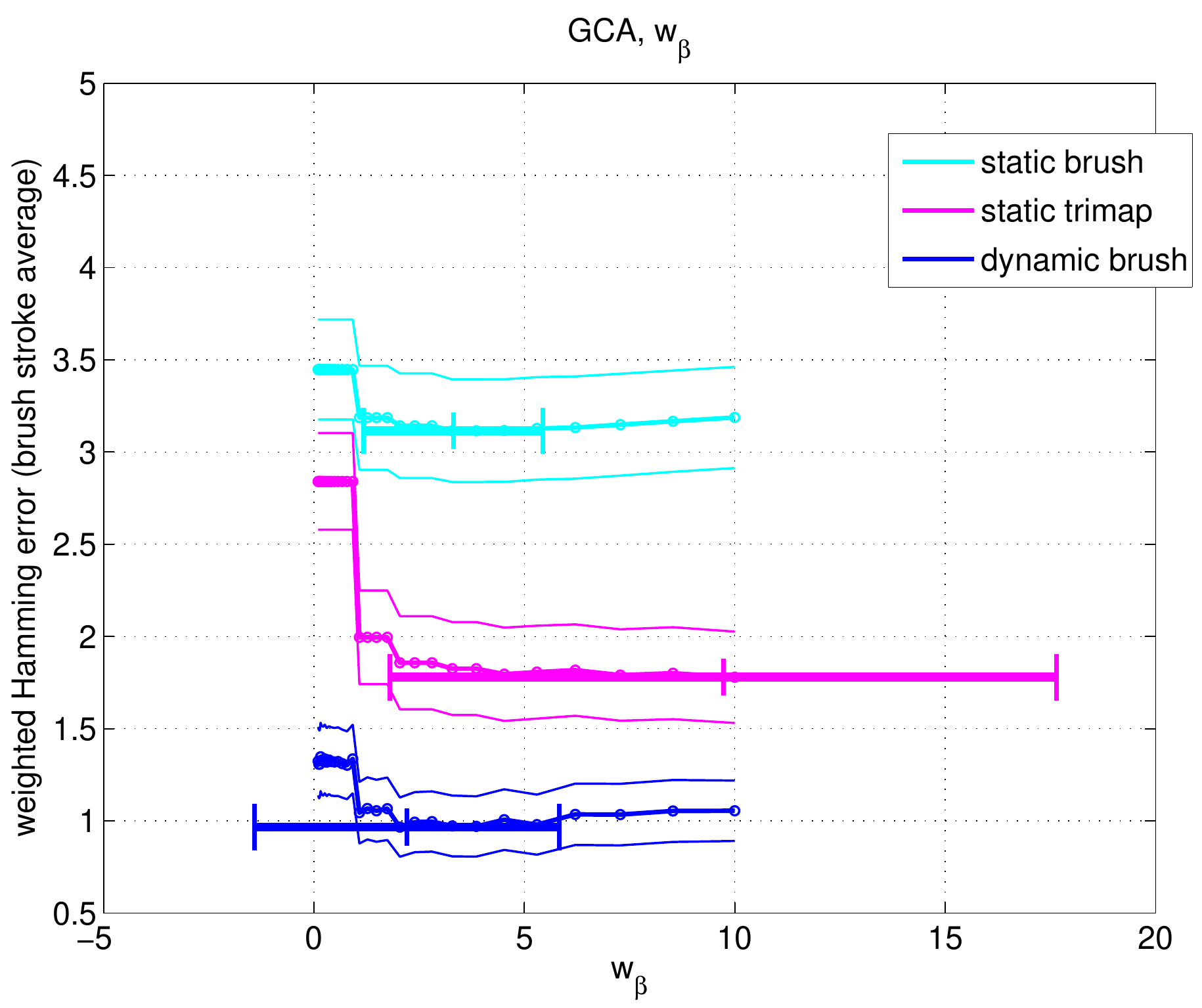}
 \par\end{centering}

 }\hfill{}\subfloat[GC, contrast weight $w_c$]{\begin{centering}
 \includegraphics[width=0.6\columnwidth]{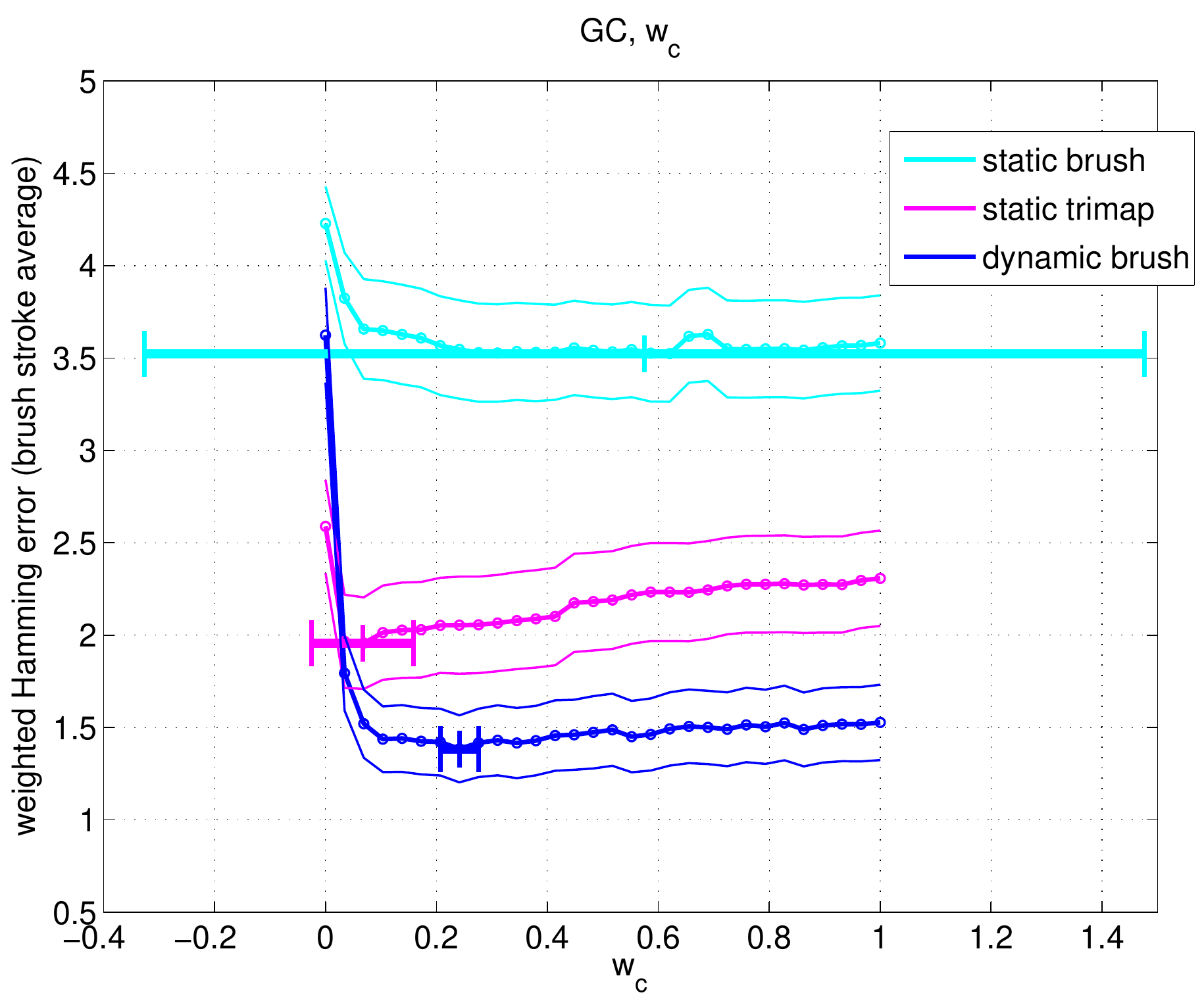}
 \par\end{centering}

 }\hfill{}\subfloat[GC, Ising weight $w_i$]{\begin{centering}
 \includegraphics[width=0.6\columnwidth]{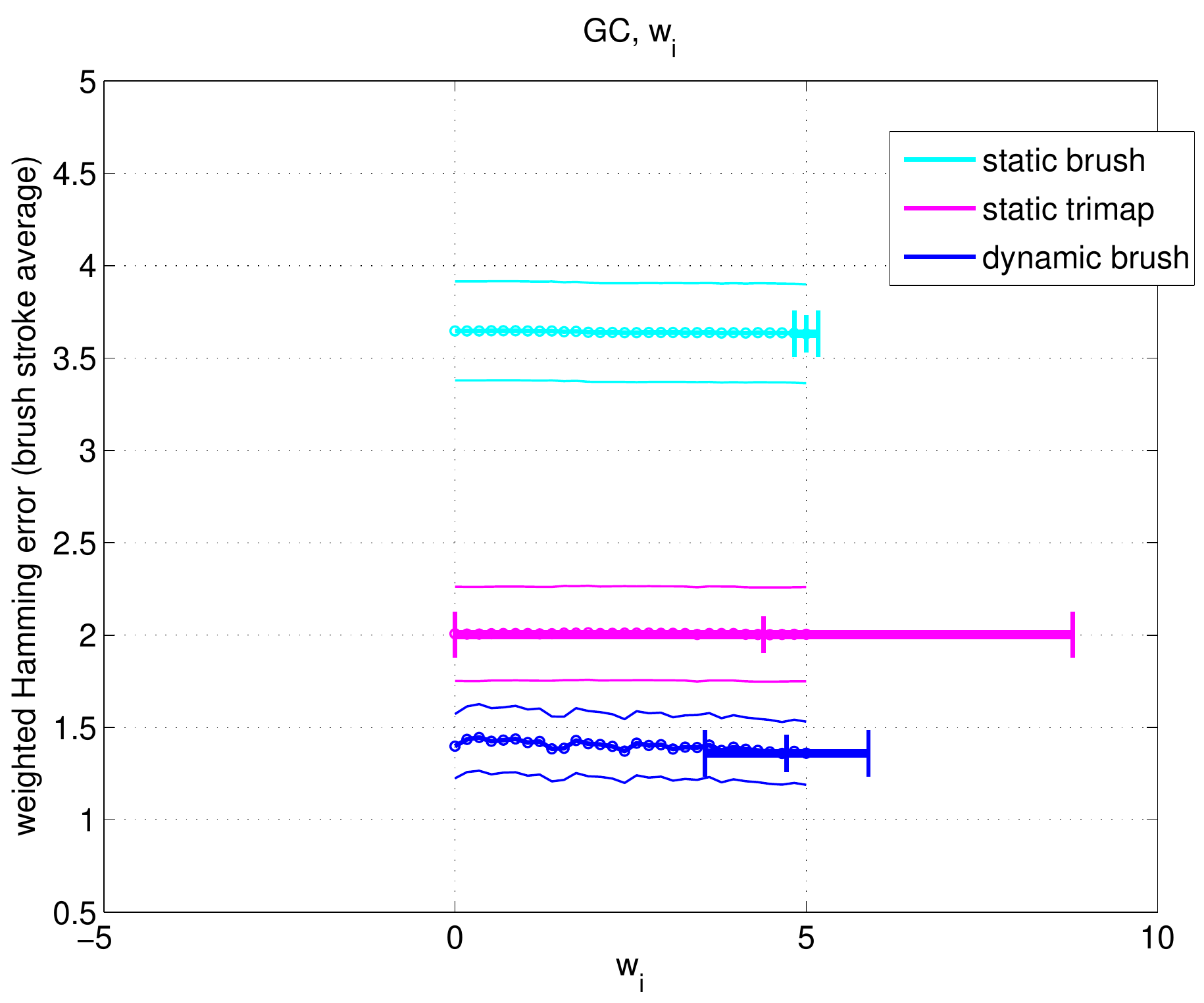}
 \par\end{centering}

 }\hfill{}\subfloat[GC, $\beta$-scale $w_\beta$]{\begin{centering}
 \includegraphics[width=0.6\columnwidth]{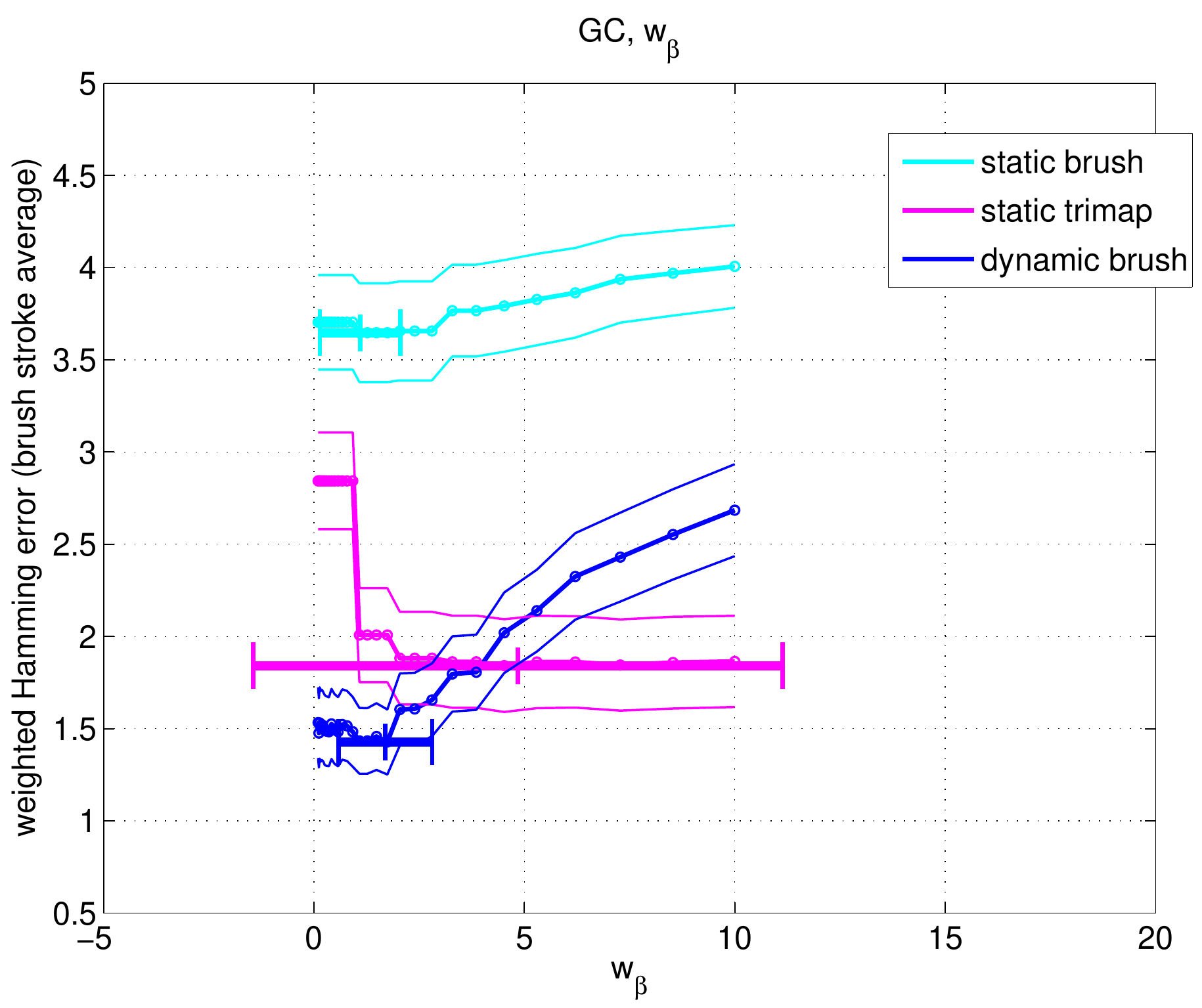}
 \par\end{centering}

 }\hfill{}\subfloat[GCA, contrast weight $w_c$]{\begin{centering}
 \includegraphics[width=0.6\columnwidth]{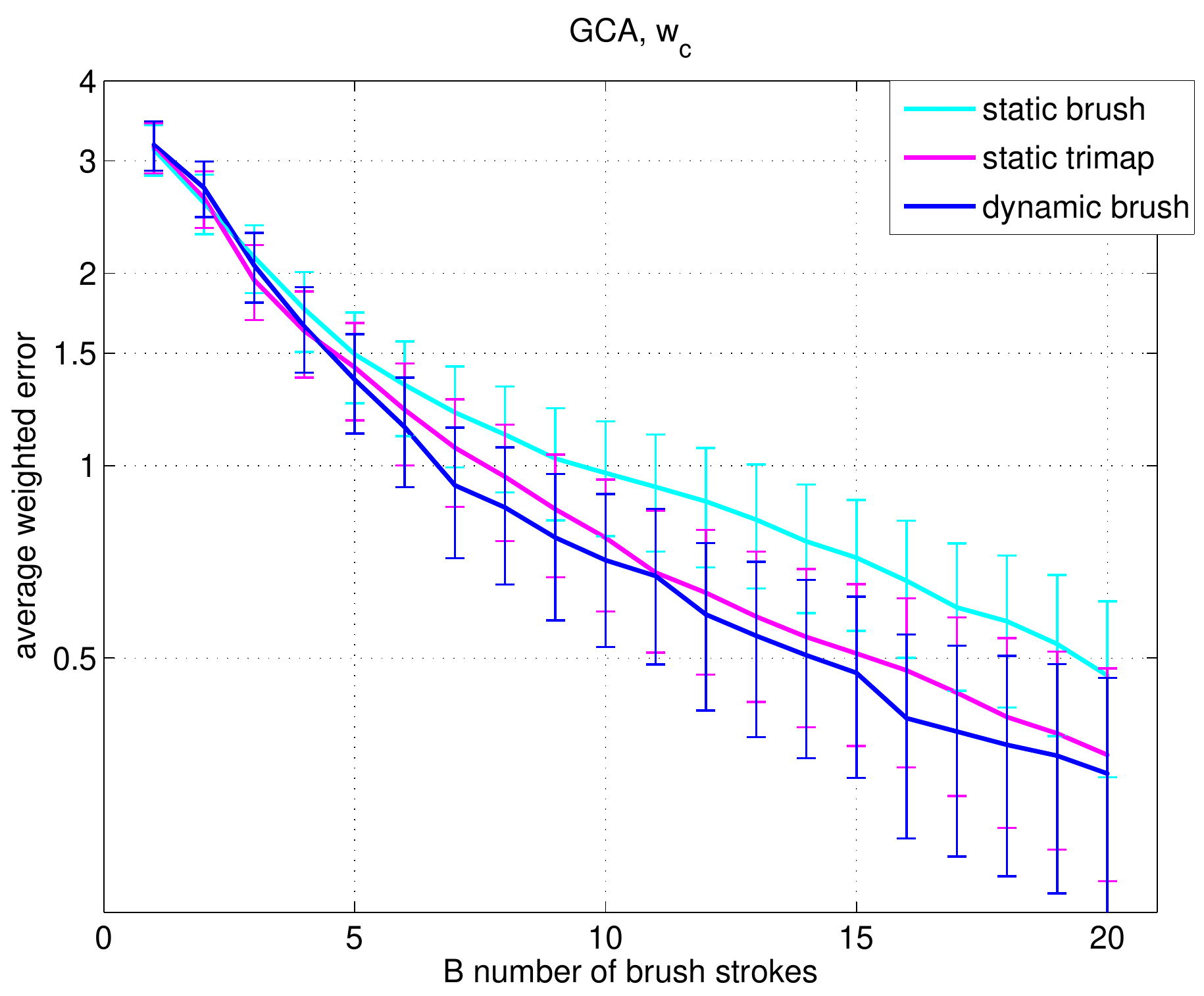}
 \par\end{centering}

 }\hfill{}\subfloat[GCA, Ising weight $w_i$]{\begin{centering}
 \includegraphics[width=0.6\columnwidth]{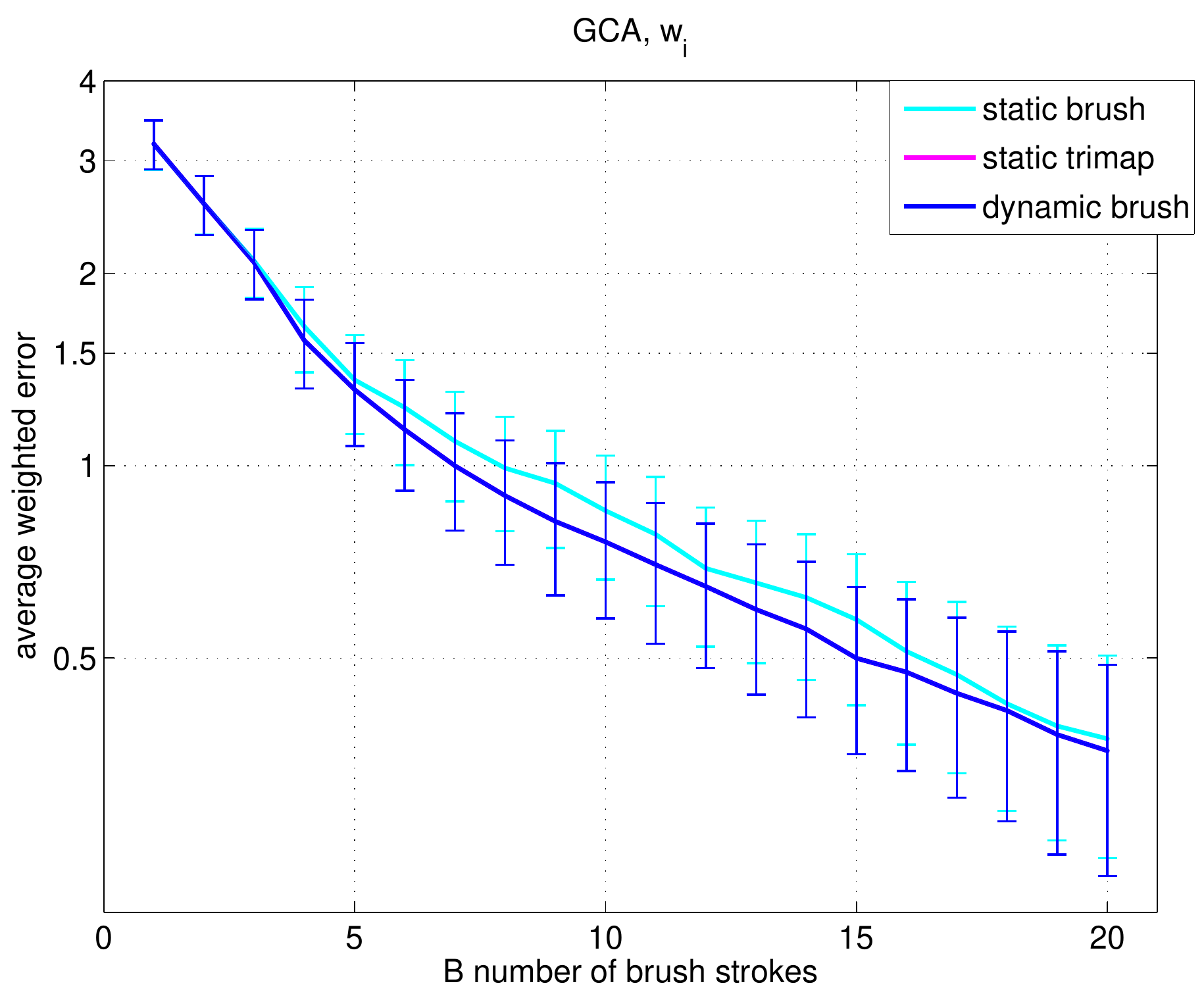}
 \par\end{centering}

 }\hfill{}\subfloat[GCA, $\beta$-scale $w_\beta$]{\begin{centering}
 \includegraphics[width=0.6\columnwidth]{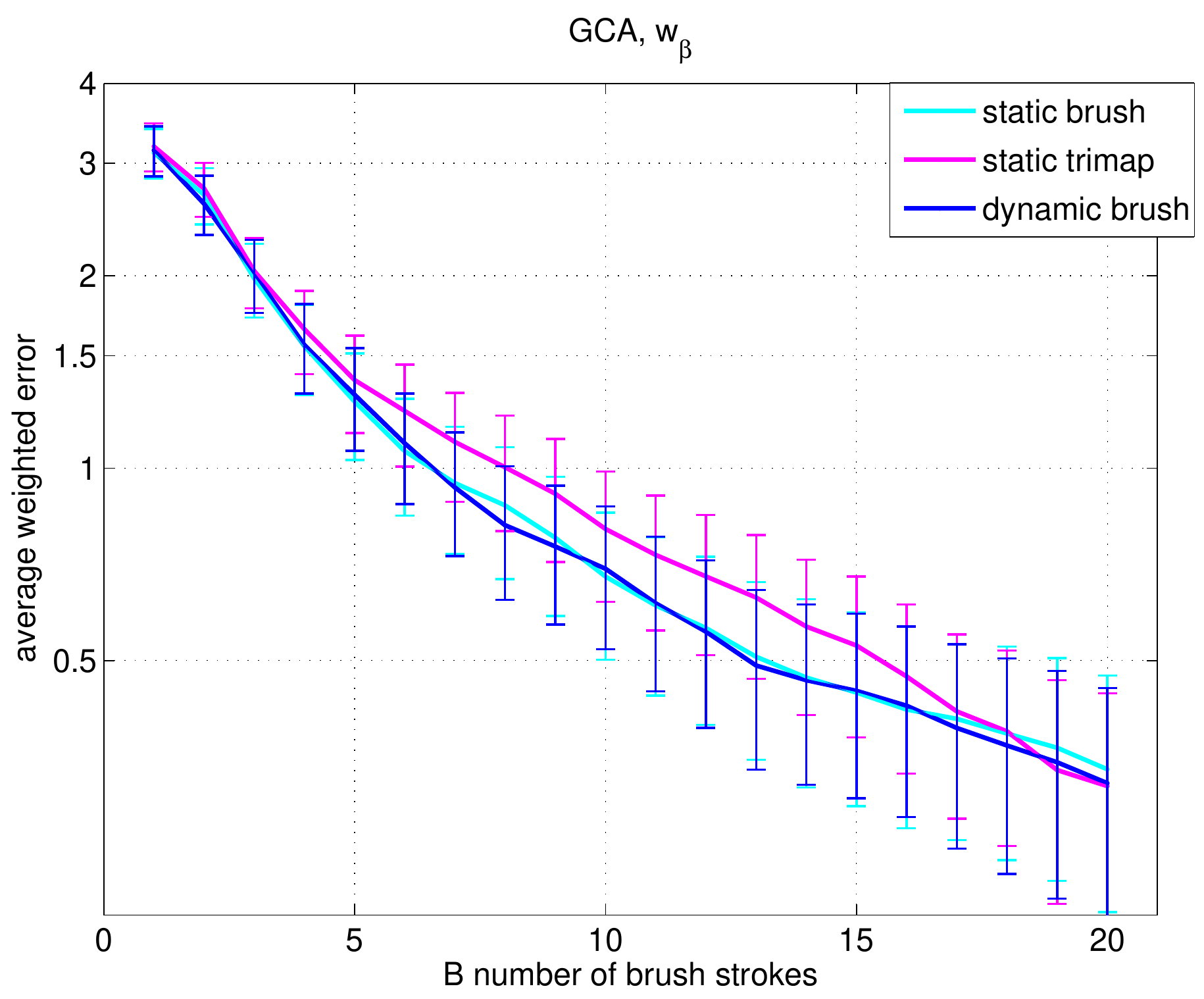}
 \par\end{centering}

 }\hfill{}\subfloat[GC, contrast weight $w_c$]{\begin{centering}
 \includegraphics[width=0.6\columnwidth]{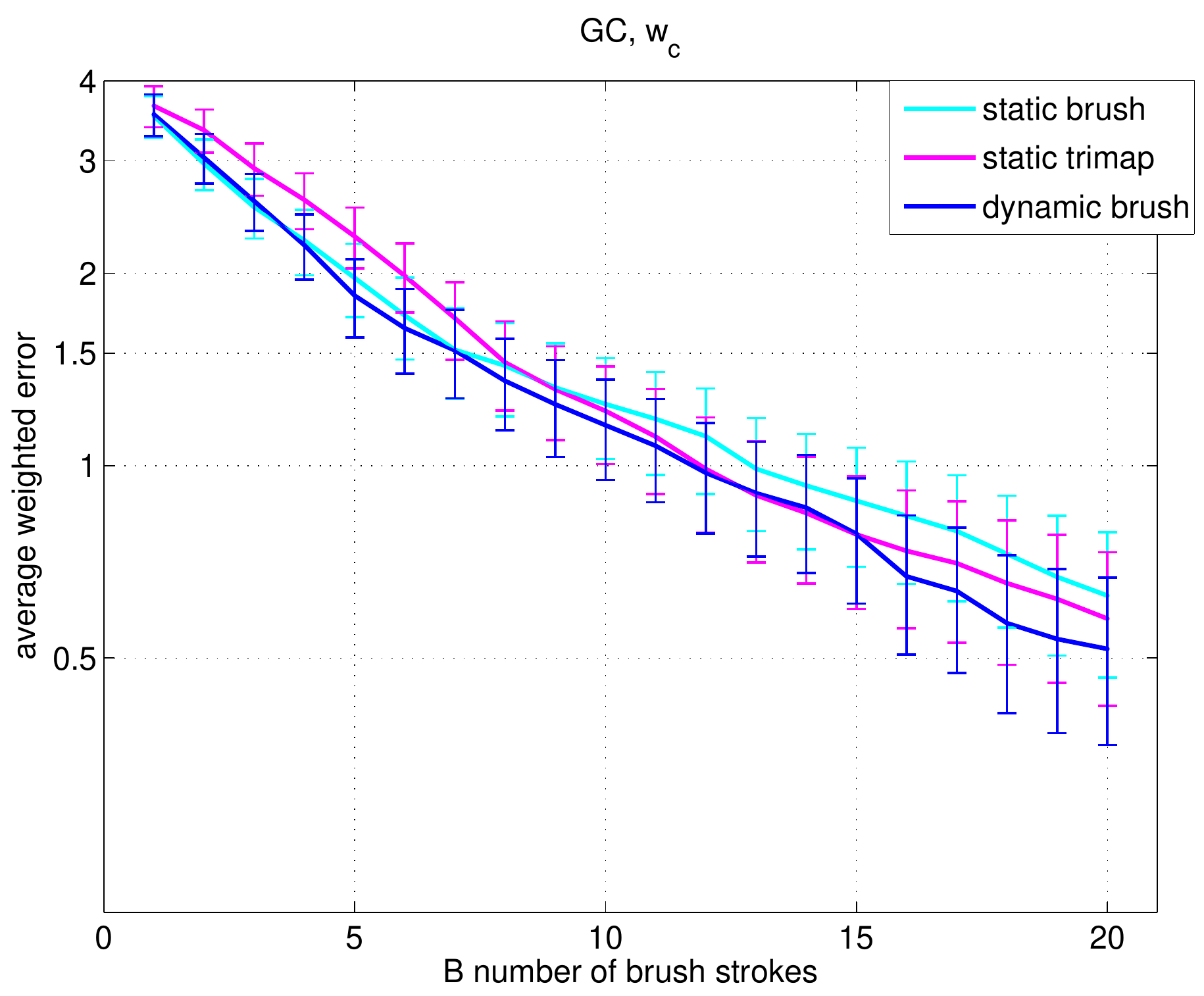}
 \par\end{centering}

 }\hfill{}\subfloat[GC, Ising weight $w_i$]{\begin{centering}
 \includegraphics[width=0.6\columnwidth]{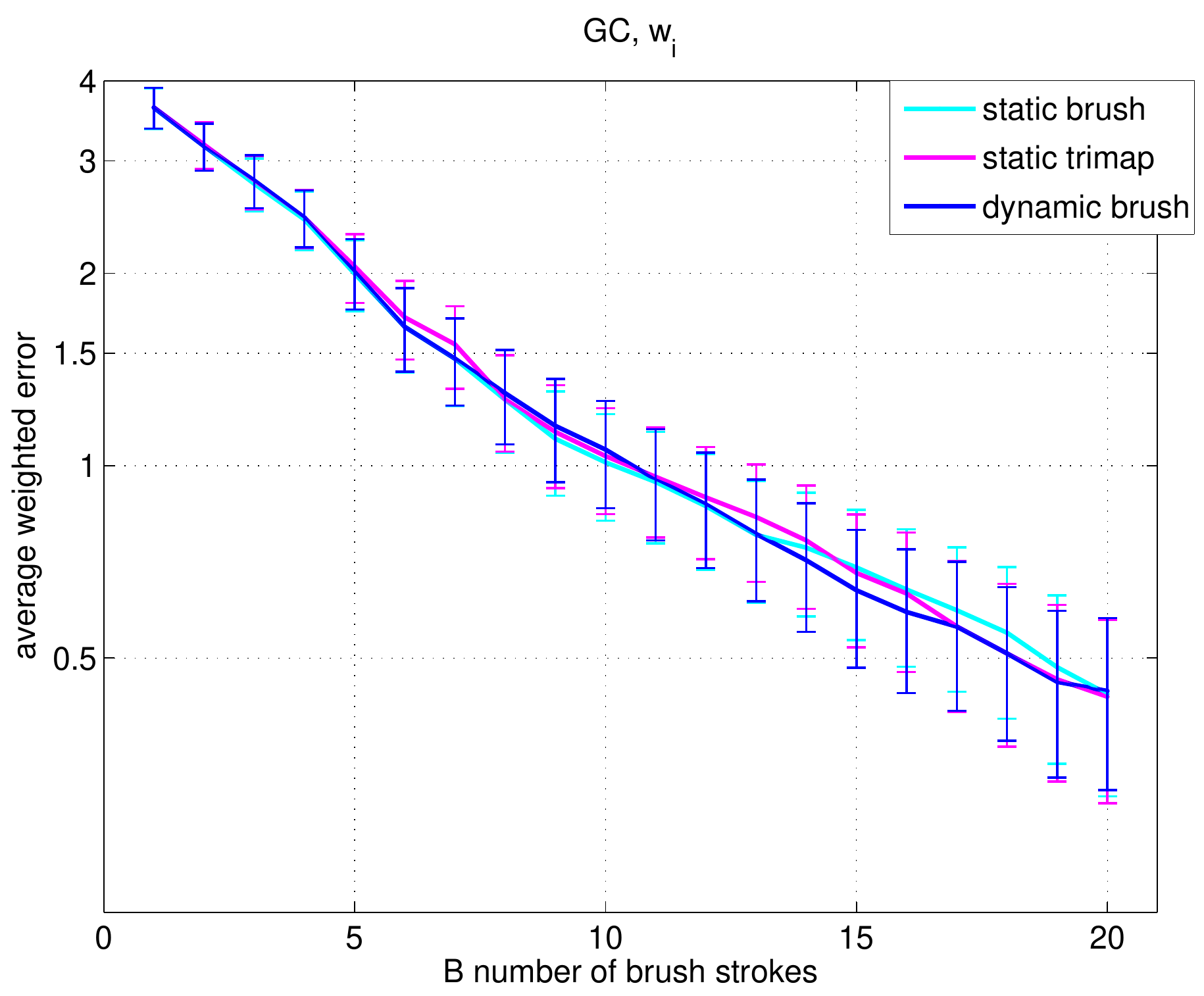}
 \par\end{centering}

 }\hfill{}\subfloat[GC, $\beta$-scale $w_\beta$]{\begin{centering}
 \includegraphics[width=0.6\columnwidth]{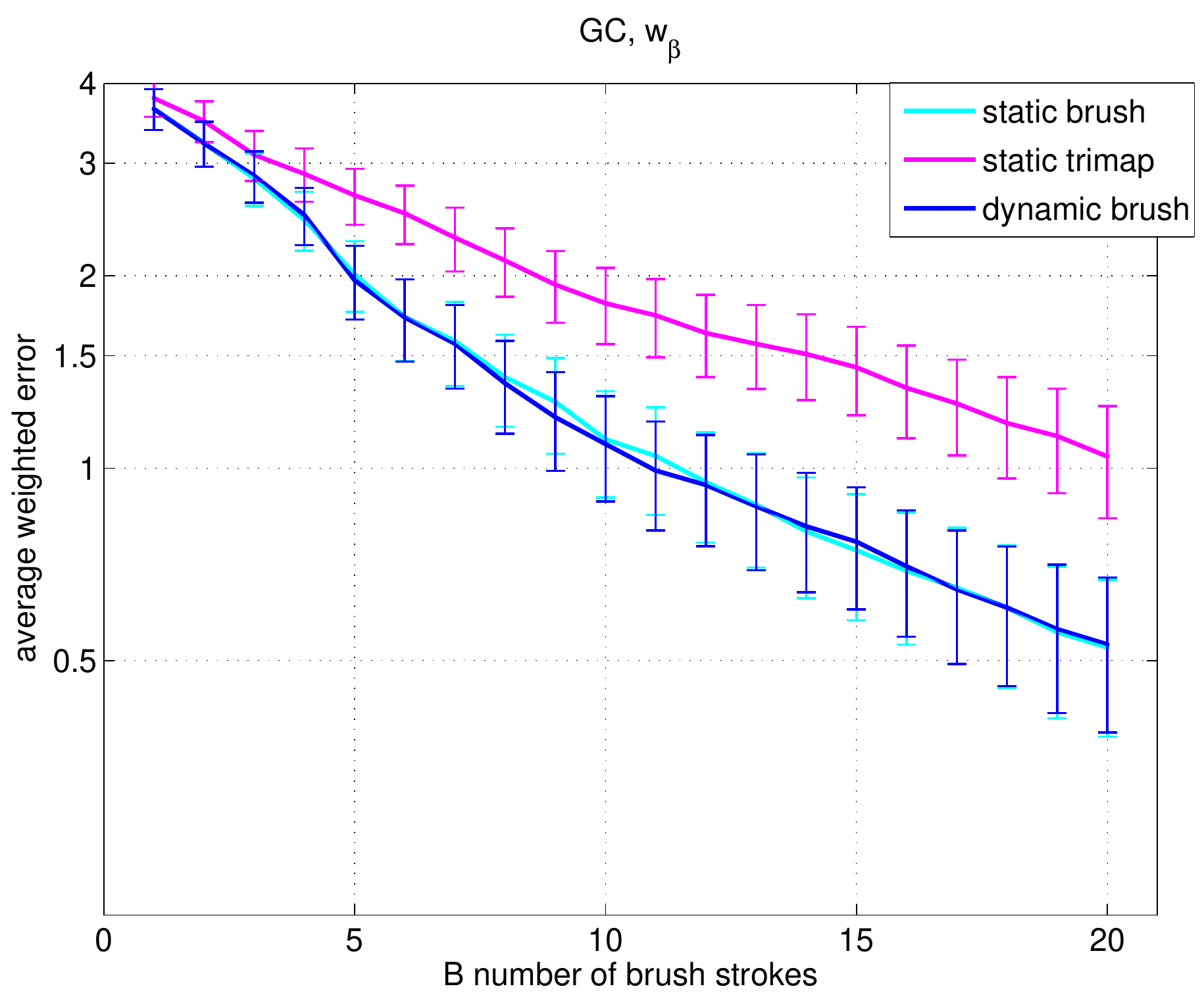}
 \par\end{centering}

}
\caption{\label{fig:gridsearch-sig} Learning with grid search (single parameter at a time), $f(er_b)=sigmoid(er_b)$, a-f training and g-l testing}
\end{figure*}

\begin{figure*}[h]
 \subfloat[GCA, contrast weight $w_c$]{\begin{centering}
 \includegraphics[width=0.6\columnwidth]{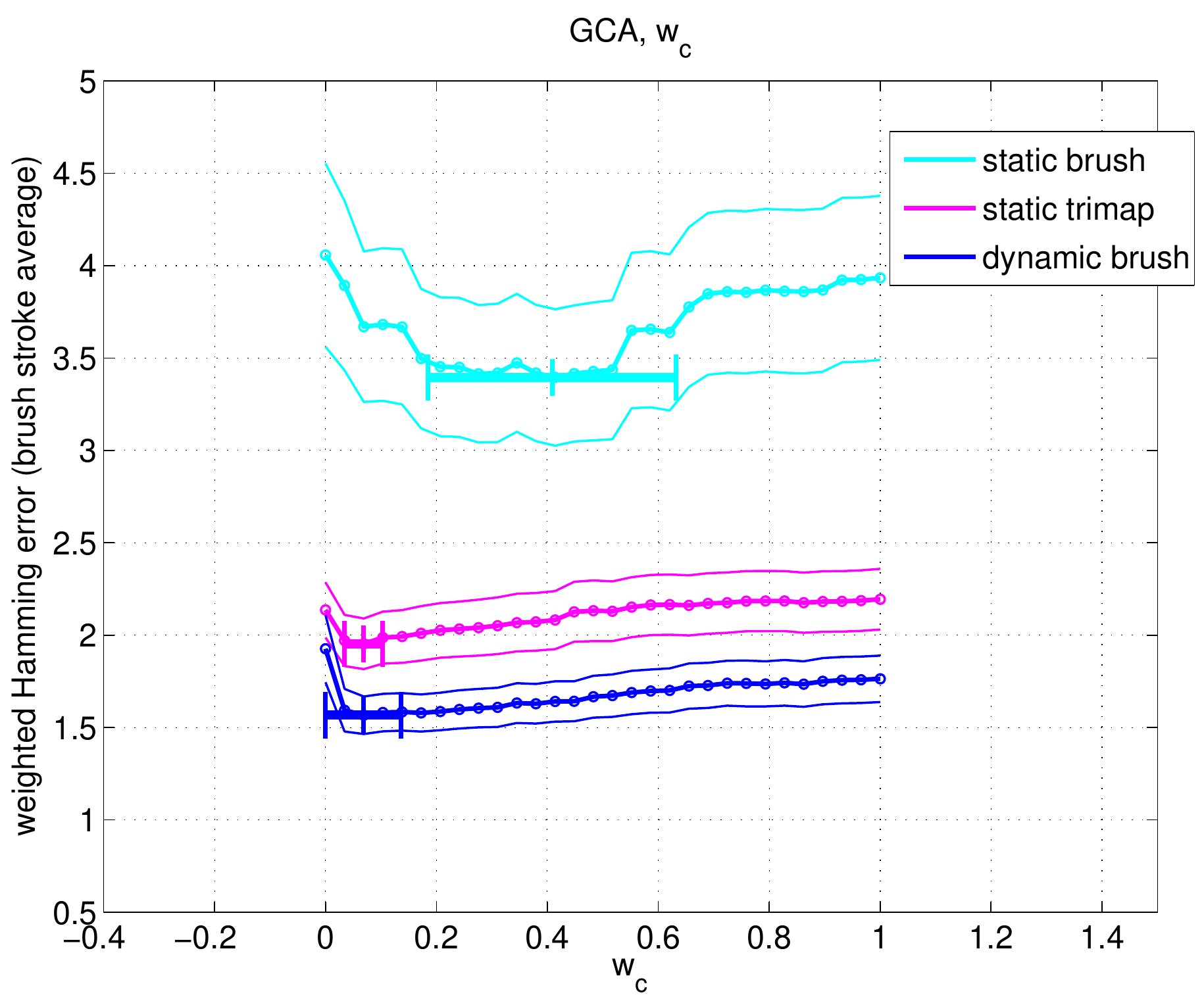}
 \par\end{centering}

 }\hfill{}\subfloat[GCA, Ising weight $w_i$]{\begin{centering}
 \includegraphics[width=0.6\columnwidth]{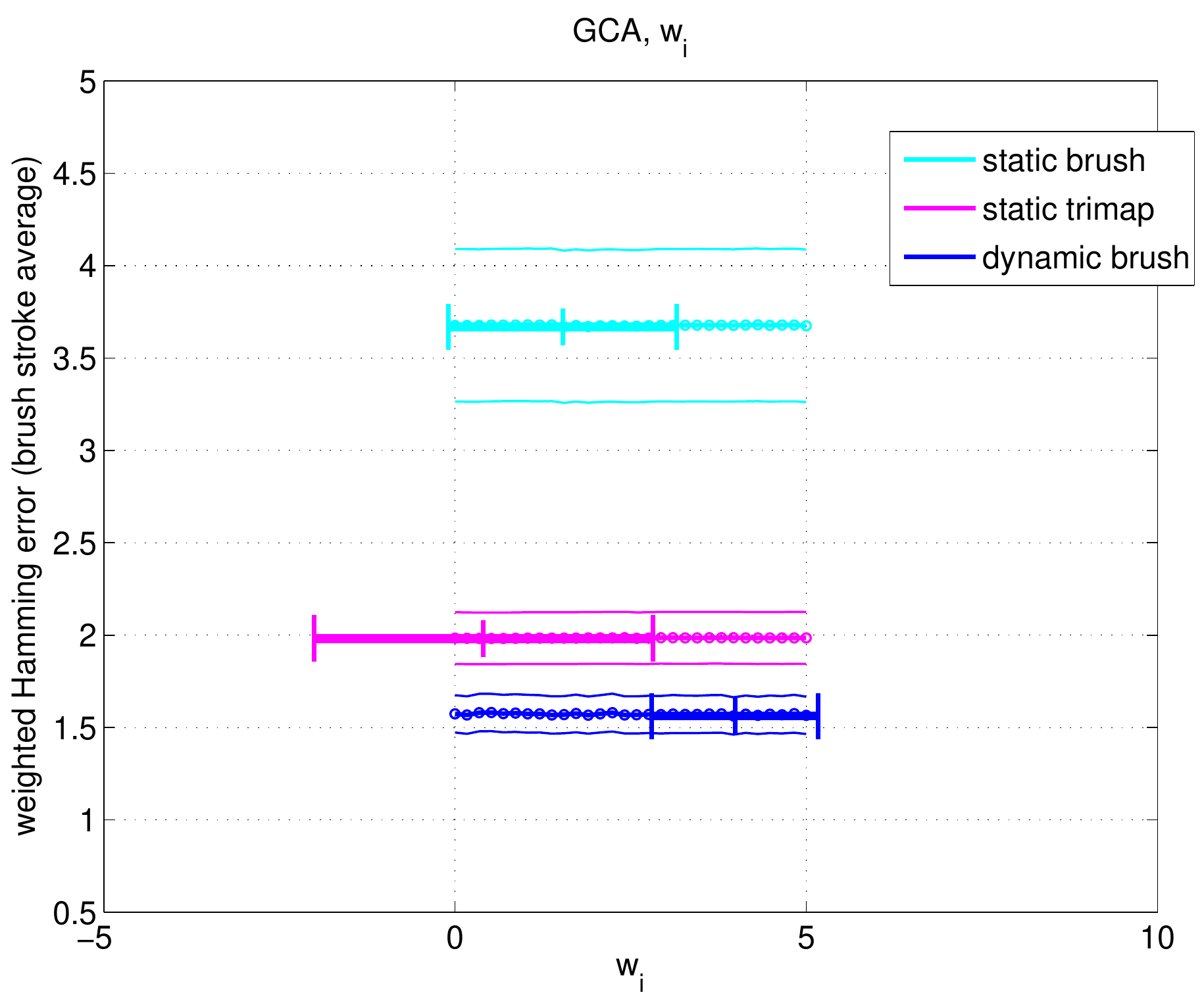}
 \par\end{centering}

 }\hfill{}\subfloat[GCA, $\beta$-scale $w_\beta$]{\begin{centering}
 \includegraphics[width=0.6\columnwidth]{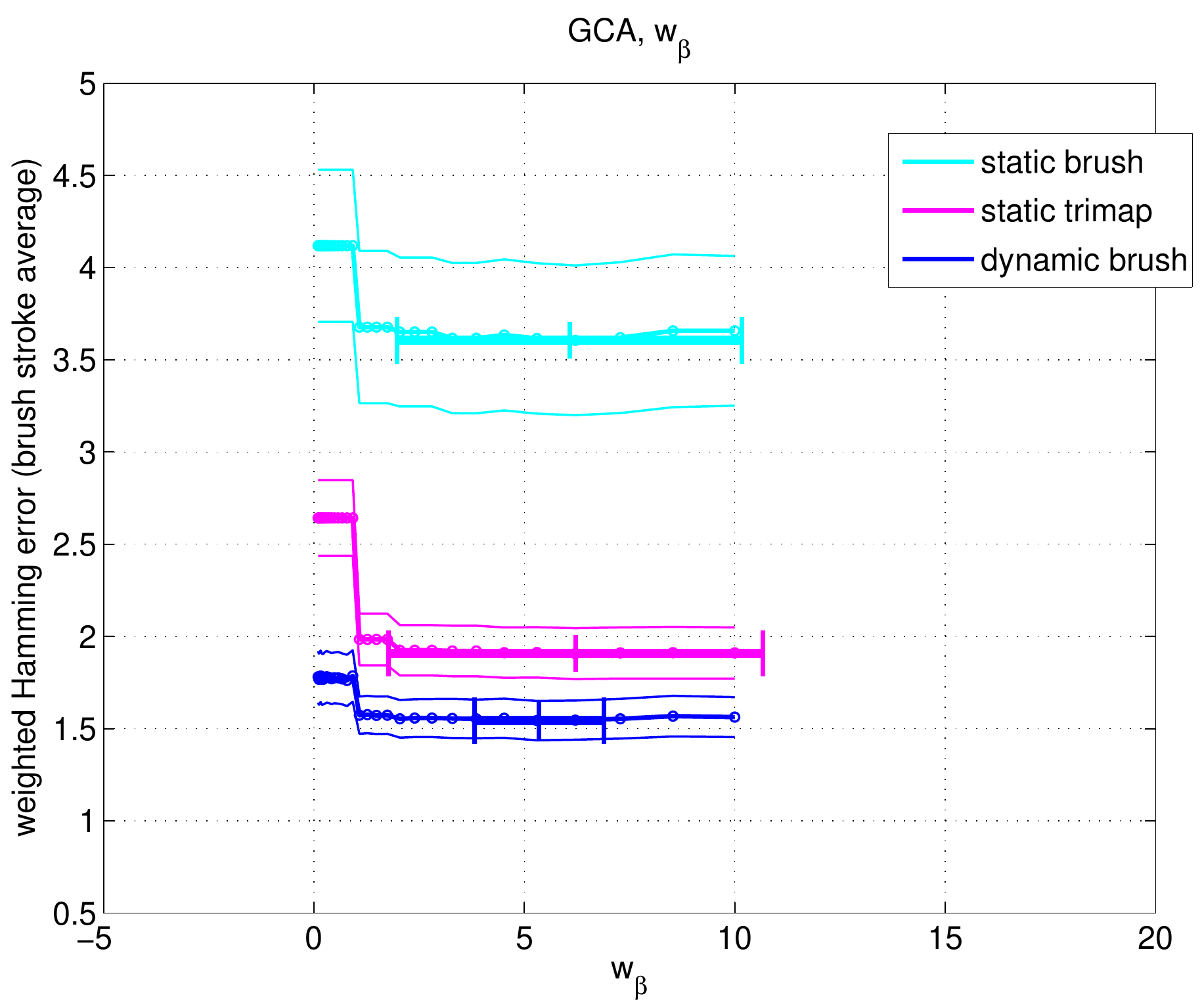}
 \par\end{centering}

 }\hfill{}\subfloat[GC, contrast weight $w_c$]{\begin{centering}
 \includegraphics[width=0.6\columnwidth]{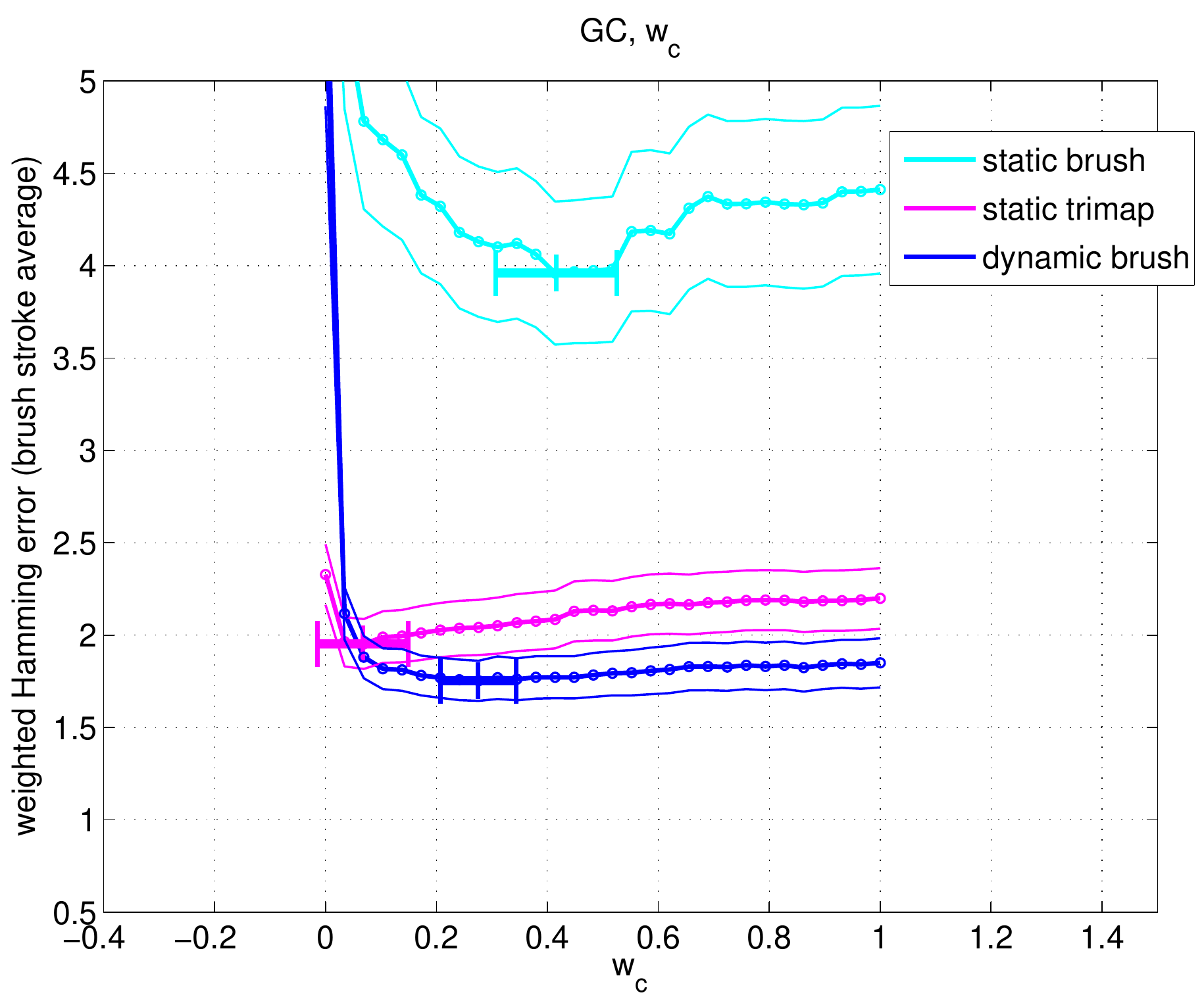}
 \par\end{centering}

 }\hfill{}\subfloat[GC, Ising weight $w_i$]{\begin{centering}
 \includegraphics[width=0.6\columnwidth]{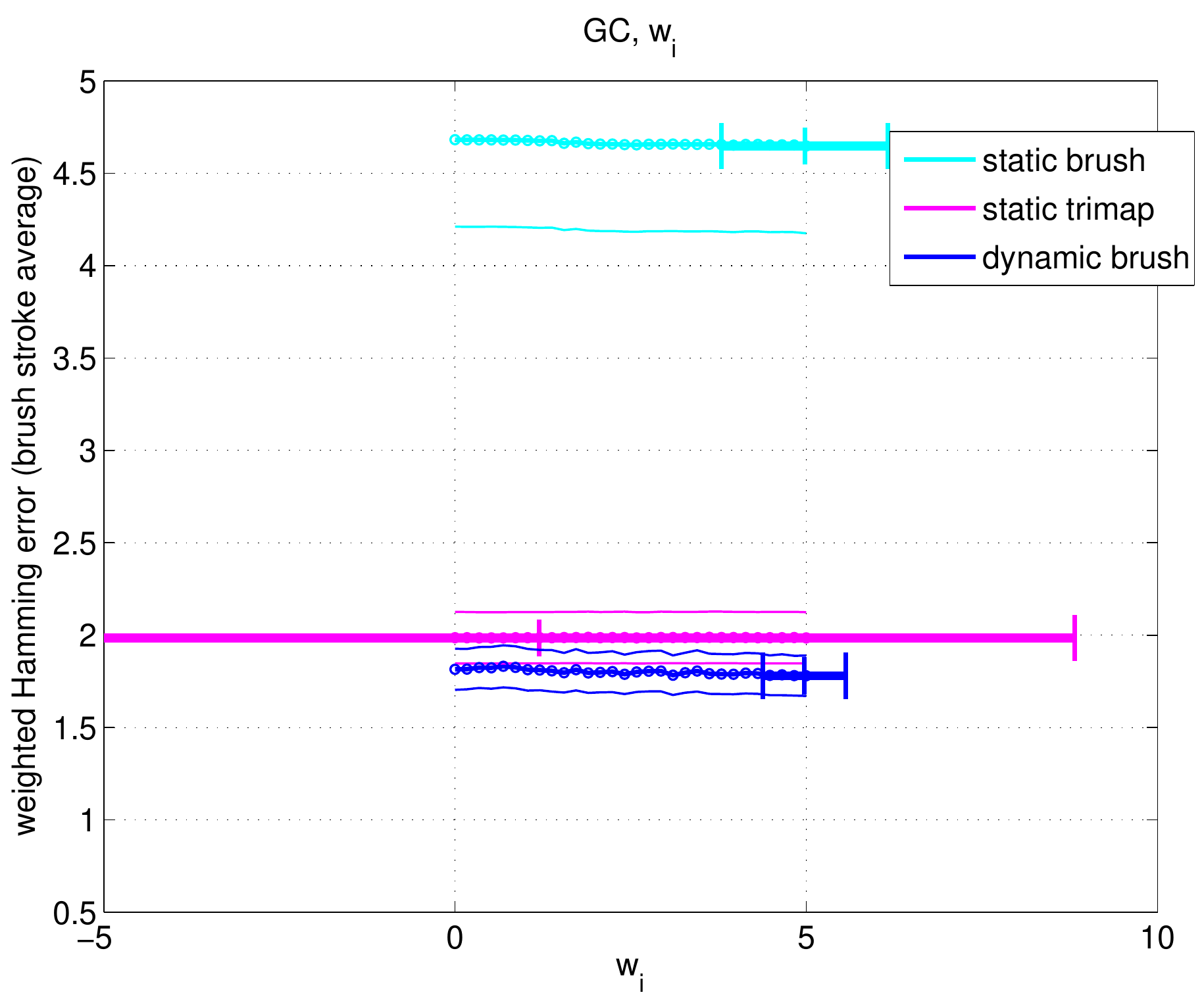}
 \par\end{centering}

 }\hfill{}\subfloat[GC, $\beta$-scale $w_\beta$]{\begin{centering}
 \includegraphics[width=0.6\columnwidth]{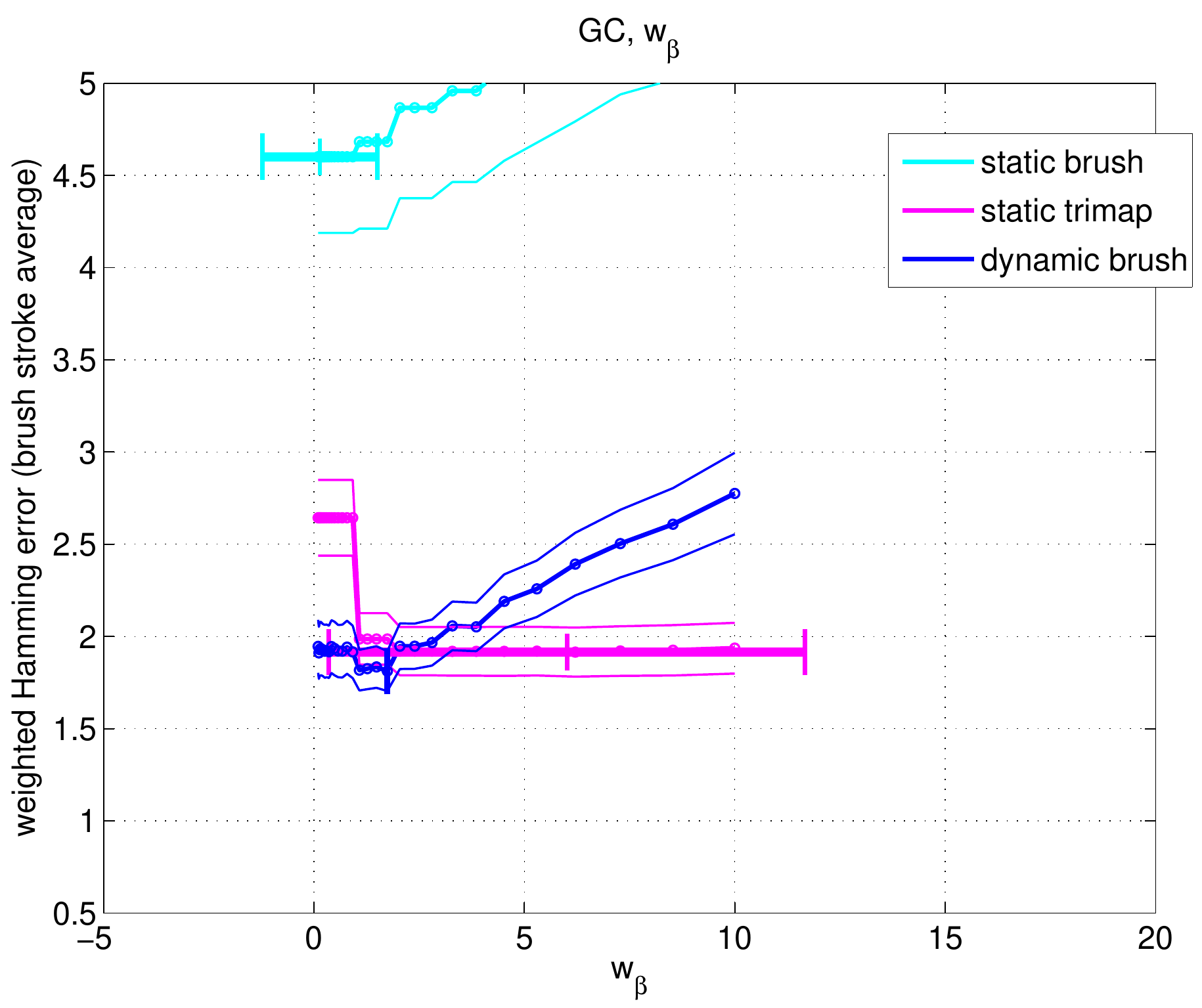}
 \par\end{centering}

 }\hfill{}\subfloat[GCA, contrast weight $w_c$]{\begin{centering}
 \includegraphics[width=0.6\columnwidth]{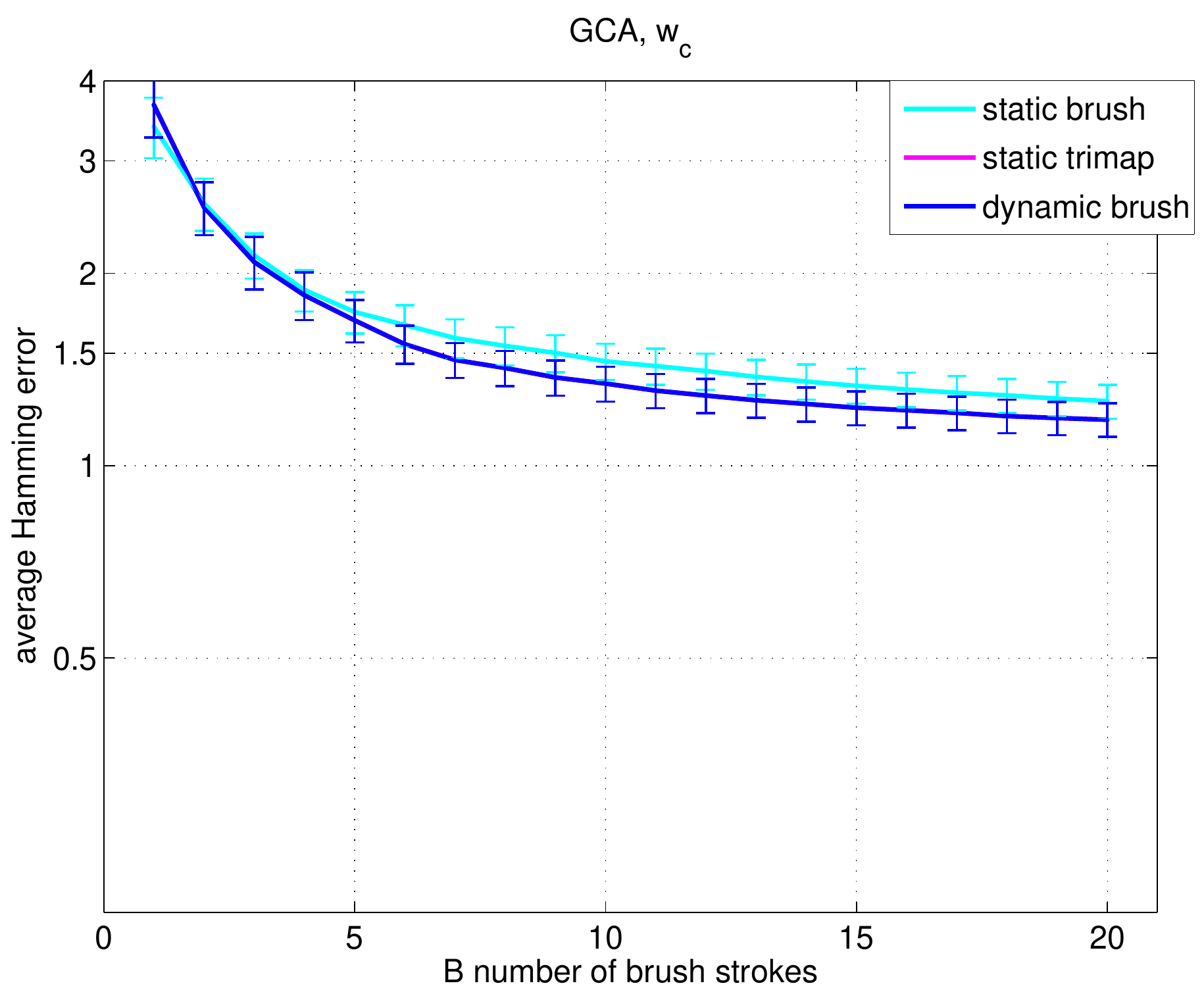}
 \par\end{centering}

 }\hfill{}\subfloat[GCA, Ising weight $w_i$]{\begin{centering}
 \includegraphics[width=0.6\columnwidth]{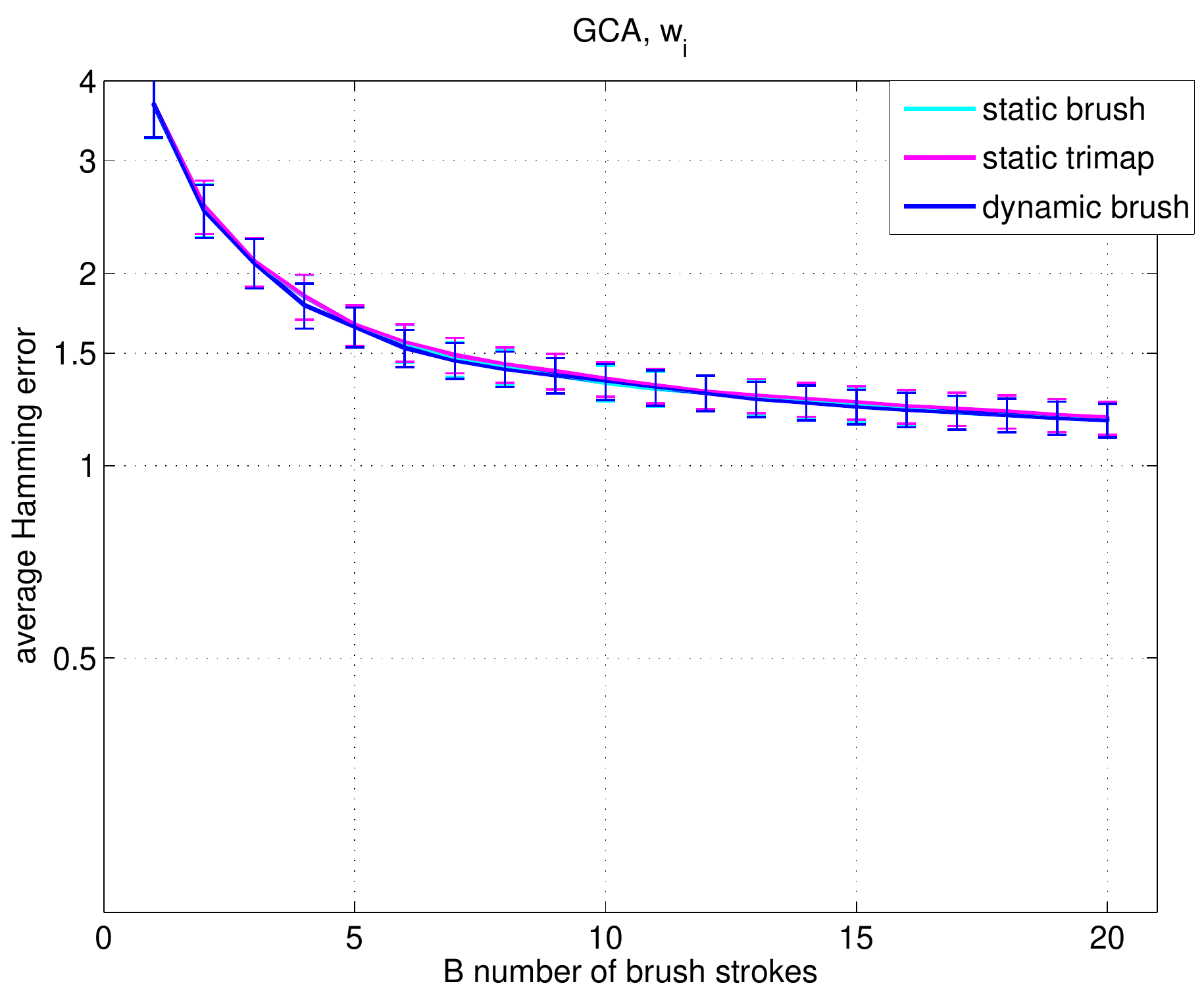}
 \par\end{centering}

 }\hfill{}\subfloat[GCA, $\beta$-scale $w_\beta$]{\begin{centering}
 \includegraphics[width=0.6\columnwidth]{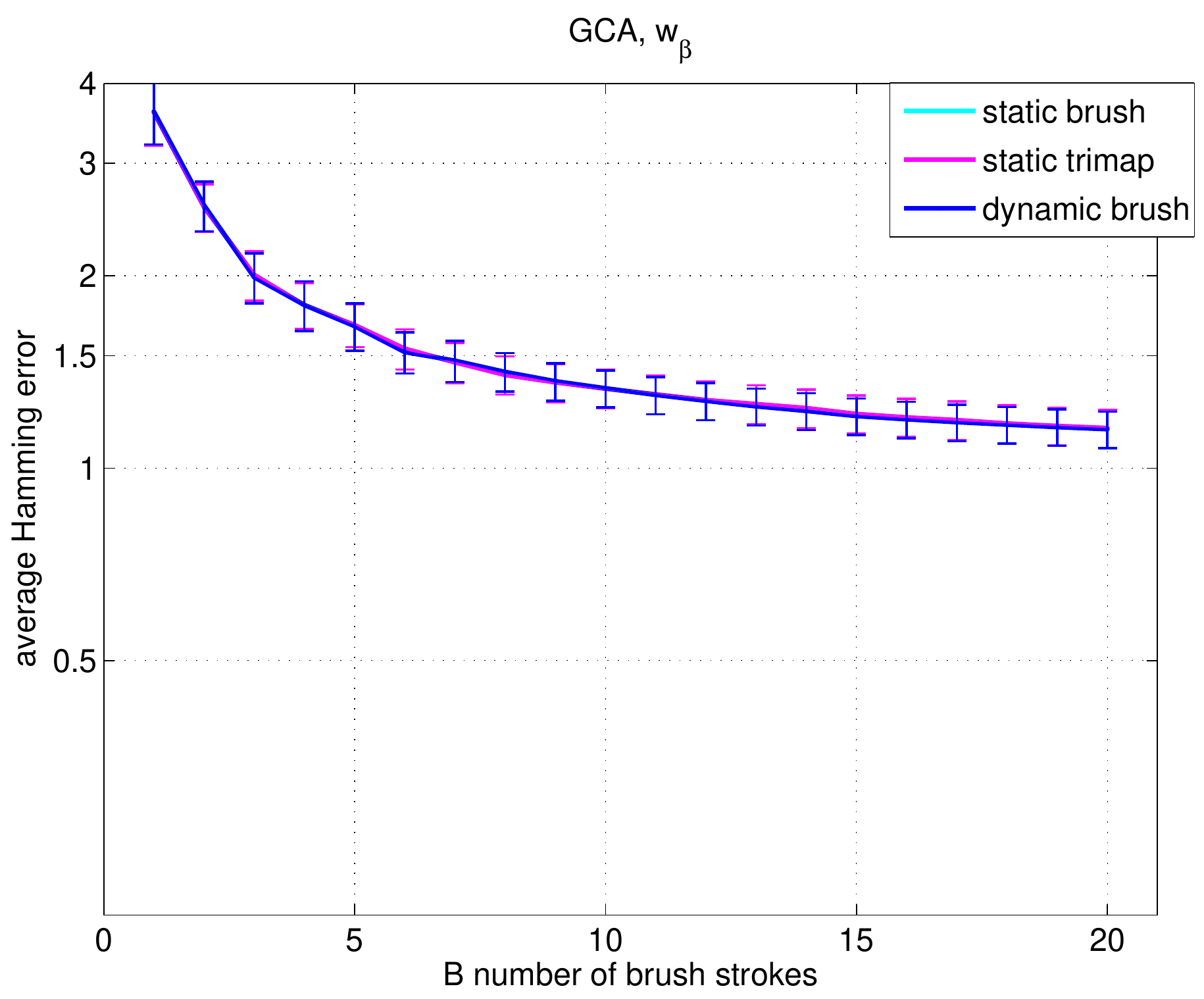}
 \par\end{centering}

 }\hfill{}\subfloat[GC, contrast weight $w_c$]{\begin{centering}
 \includegraphics[width=0.6\columnwidth]{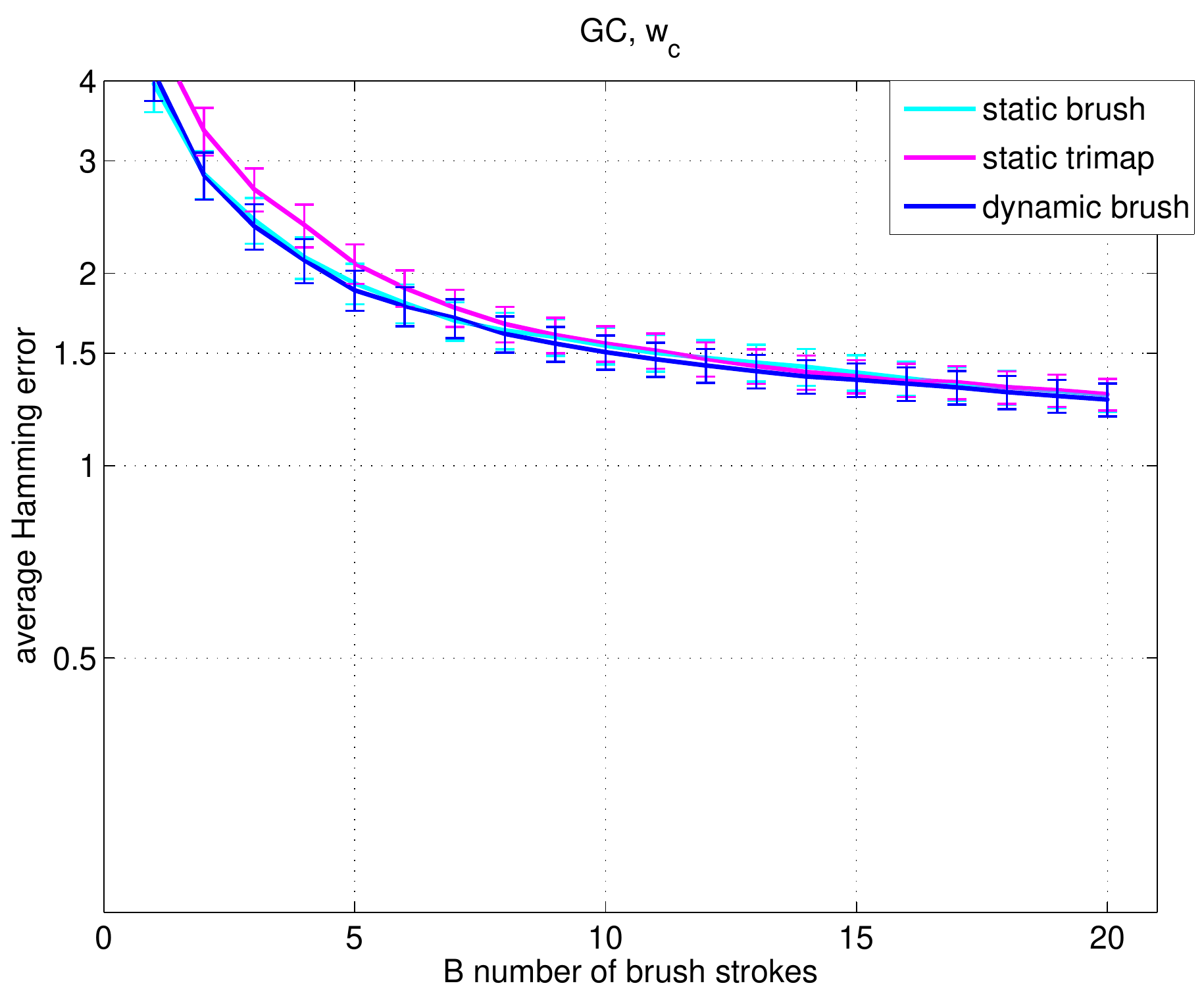}
 \par\end{centering}

 }\hfill{}\subfloat[GC, Ising weight $w_i$]{\begin{centering}
 \includegraphics[width=0.6\columnwidth]{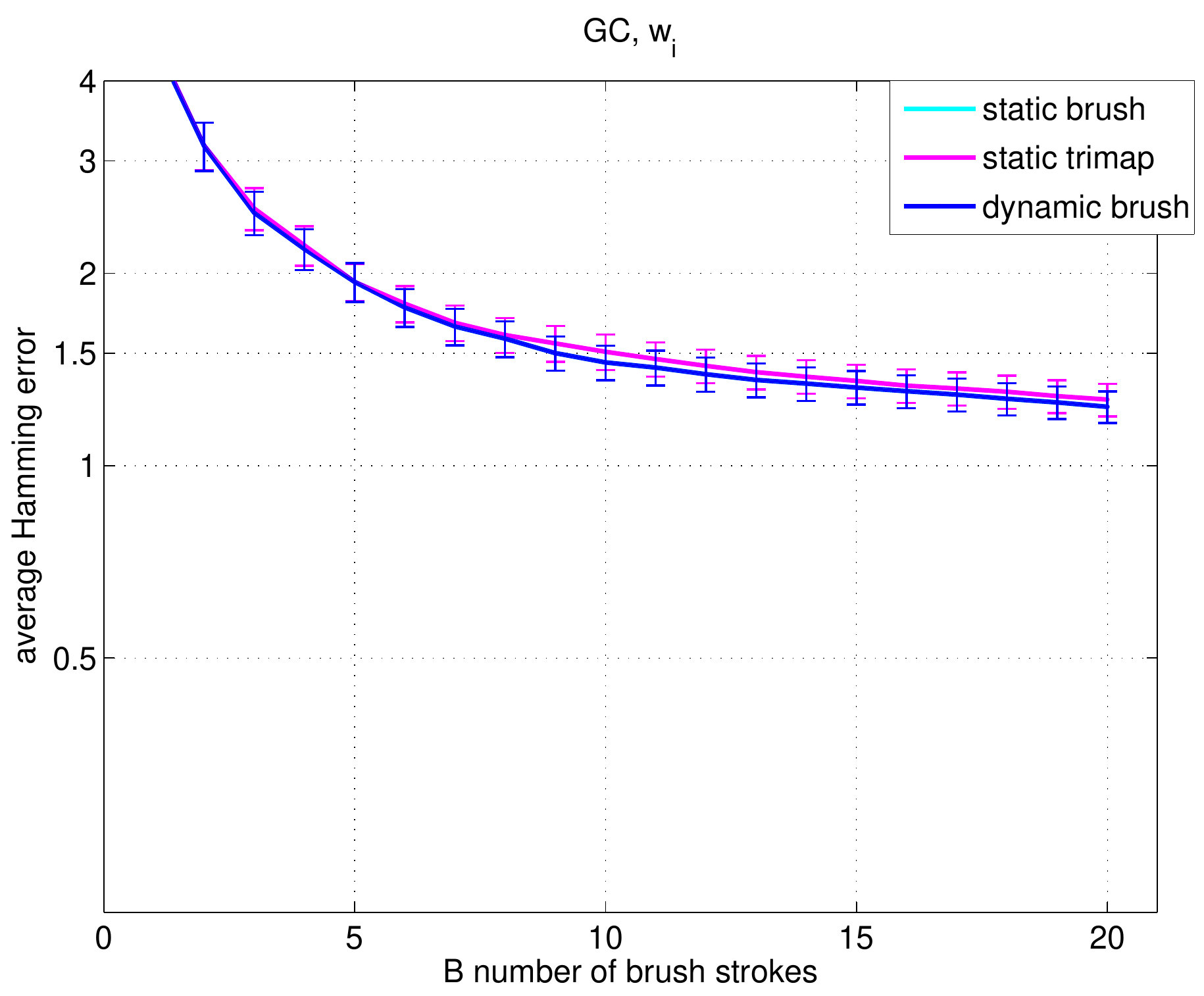}
 \par\end{centering}

 }\hfill{}\subfloat[GC, $\beta$-scale $w_\beta$]{\begin{centering}
 \includegraphics[width=0.6\columnwidth]{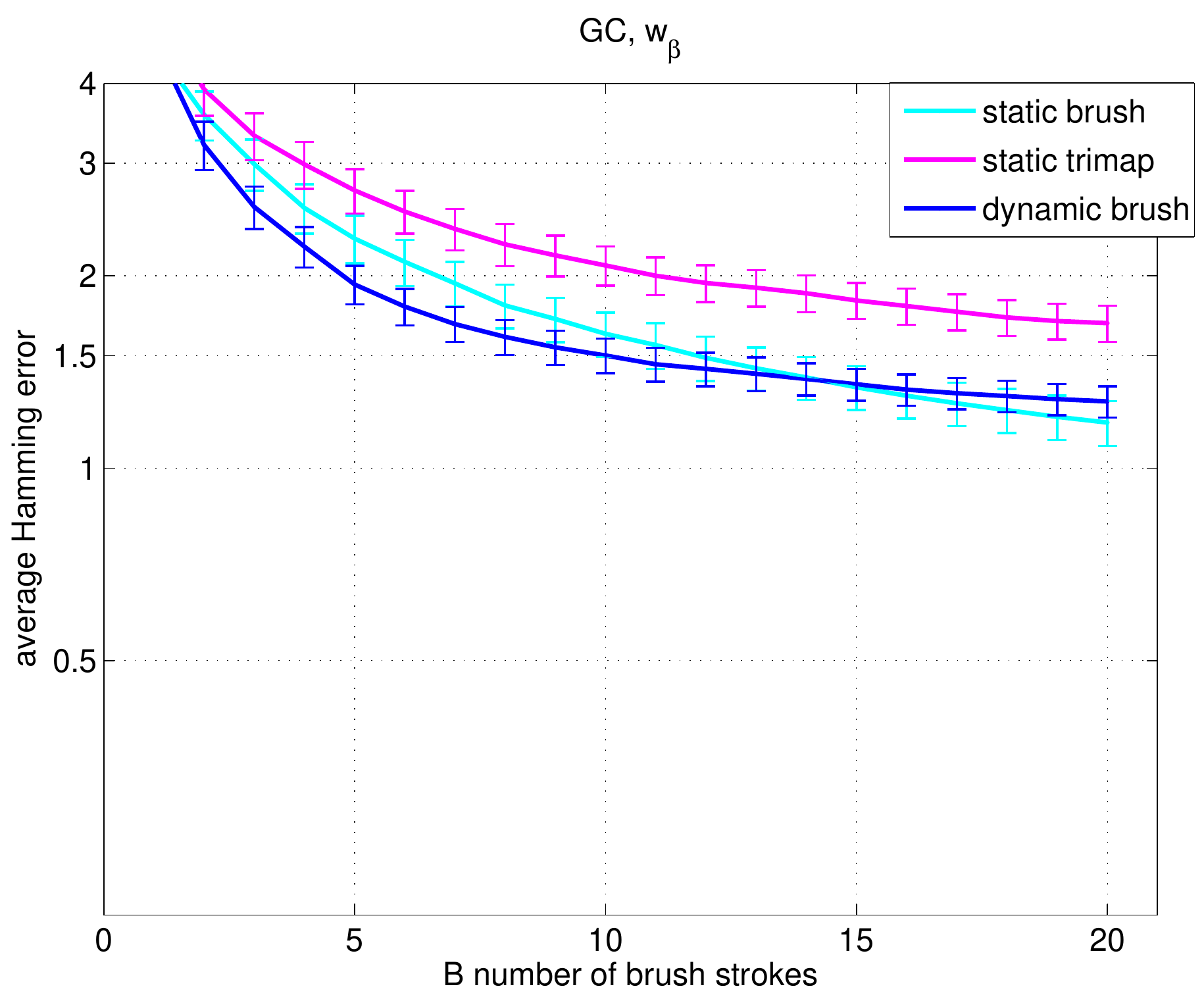}
 \par\end{centering}

}
\caption{\label{fig:gridsearch-id} Learning with grid search (single parameter at a time), $f(er_b)=er_b$, a-f training and g-l testing}
\end{figure*}

In order to get a complete picture, we provide the full set of plots for the
line search experiments. We report results for the two systems GCA and GC on
three parameters $w_c$, $w_i$ and $w_\beta$ and two error weighting functions
$f(er_b)$ in Figures \ref{fig:gridsearch-sig} and \ref{fig:gridsearch-id}.

\paragraph{Novice vs Advanced User} When comparing different
interactive systems, we have to decide, whether the system 
is designed for an advanced or a novice user.

In a user study, one has full control over selecting advanced or
novice users. This can be done by changing the amount of
introduction given to the participants. However, this process
is lengthy and therefore infeasible for learning.

\begin{figure}[h]
\begin{centering}
\includegraphics[width=0.95\columnwidth]{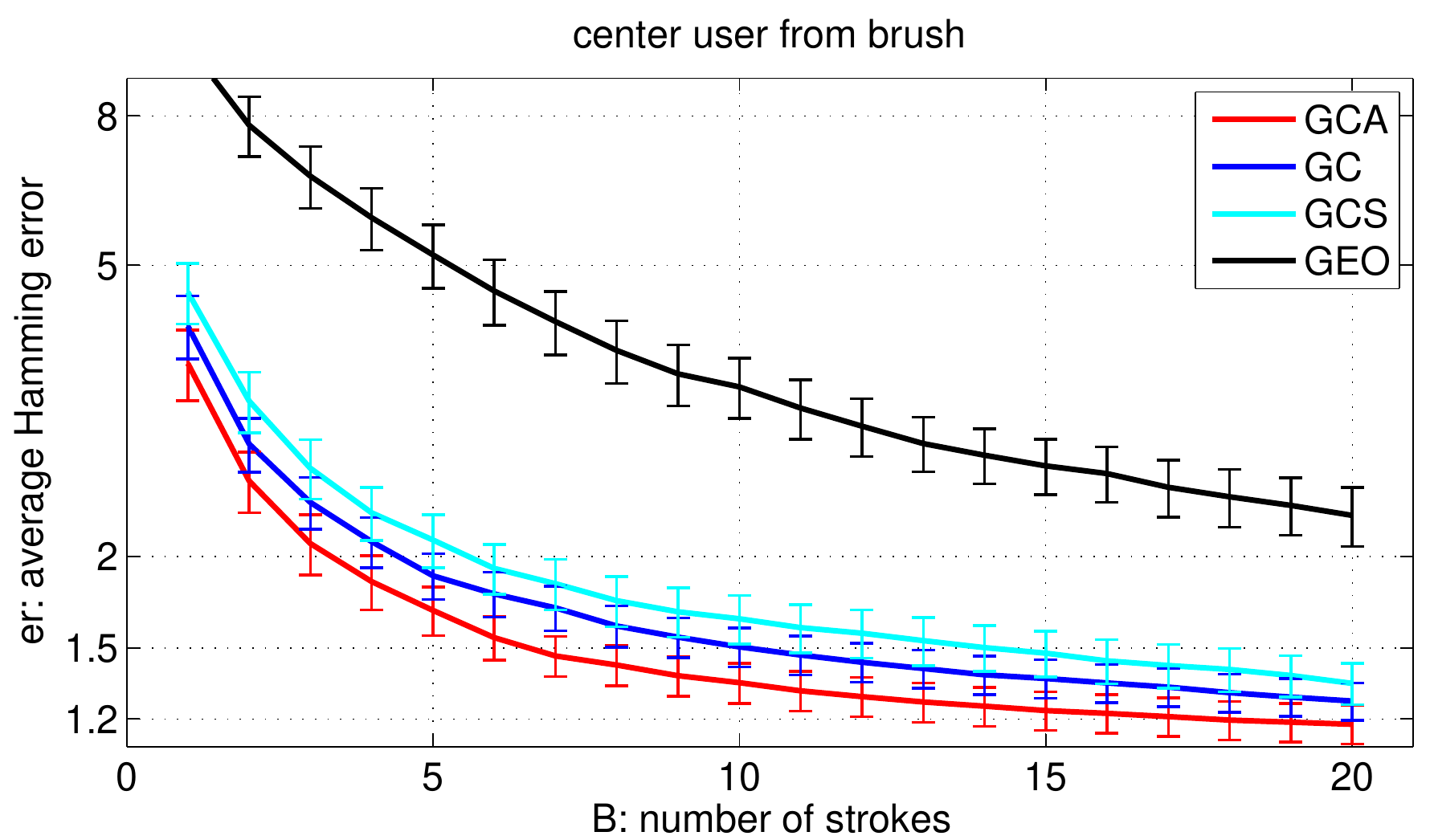}
\par\end{centering}

\caption{\label{fig:systemcompare}{\it System comparison}: Segmentation performance comparison between 4 different systems: GCA, GC, GCS and GEO using the robot user started from initial user brushes.}
\end{figure}

In our robot user paradigm, we can simulate users with
different levels of experience. We run the (center) user model to
simulate a novice user and evaluate four different systems. The
results are shown in Fig. \ref{fig:systemcompare}. The order of the
methods is as expected, GCA is best, followed by GC, then GCS, and
GEO. GEO performs badly since it does no smoothing at the boundary,
compared to the other systems.

\section{Interactive Max-margin Learning} \label{sec:maxmargin} The
grid-search method used in Section \ref{sec:linesearch} can be used
for learning models with few parameters only. Max-margin methods
deal which models containing large numbers of parameters
and have been used extensively in computer vision. However, they work
with static training data and cannot be used with an active user
model. In this Section, we show how the traditional max-margin
parameter learning algorithm can be extended to incorporate an active
user.

\subsection{Static SVMstruct} Our exposition builds heavily on
\cite{szummer08CRFgraphCut} and the references therein. The SVMstruct
framework \cite{tsochantaridis04SVMstruct} allows to adjust linear
parameters $\mathbf{w}$ of the segmentation energy
$E_{\mathbf{w}}(\mathbf{y},\mathbf{x})$ (Eq. \ref{eq:E}) from a
given training set $\{\mathbf{x}^{k},\mathbf{y}^{k}\}_{k=1..K}$ of
$K$ images $\mathbf{x}^{k}\in\mathbb{R}^{n}$ and ground truth
segmentations%
\footnote{We write images of size $(n_{x},n_{y},n_{c})$ as vectors for
simplicity. All involved operations respect the $2$d grid
structure absent in general $n$-vectors.%
} $\mathbf{y}\in\mathcal{Y}:=\{0,1\}^{n}$ by balancing between empirical
risk $\sum_{k}\Delta(\mathbf{y}^{k},f(\mathbf{x}^{k}))$ and regularisation
by means of a trade-off parameter $C$. A (symmetric) loss function%
\footnote{We use the Hamming loss $\Delta_{H}(\mathbf{y}^{*},\mathbf{y}^{k})=\mathbf{1}^{\top}|\mathbf{y}^{k}-\mathbf{y}^{*}|$.%
} $\Delta:\mathcal{Y}\times\mathcal{Y}\rightarrow\mathbb{R}_{+}$ measures
the degree of fit between two segmentations $\mathbf{y}$ and $\mathbf{y}^{*}$.
The current segmentation is given by $\mathbf{y}^{*}=\arg\min_{\mathbf{y}}E_{\mathbf{w}}(\mathbf{y},\mathbf{x})$.
We can write the energy function as an inner product between feature
functions $\psi_{i}(\mathbf{y},\mathbf{x})$ and our parameter vector
$\mathbf{w}$: $E_{\mathbf{w}}(\mathbf{y},\mathbf{x})=\mathbf{w}^{\top}\bm{\psi}(\mathbf{y},\mathbf{x})$.
With the two shortcuts $\delta\bm{\psi}_{\mathbf{y}}^{k}=\bm{\psi}(\mathbf{x}^{k},\mathbf{y})-\bm{\psi}(\mathbf{x}^{k},\mathbf{y}^{k})$
and $\ell_{\mathbf{y}}^{k}=\Delta(\mathbf{y},\mathbf{y}^{k})$, the
margin rescaled objective \cite{taskar04AMN} reads

\begin{eqnarray}
\min_{\bm{\xi}\ge\mathbf{0},\mathbf{w}} & o(\mathbf{w}):=\frac{1}{2}\left\Vert \mathbf{w}\right\Vert ^{2}+\frac{C}{K}\mathbf{1}^{\top}\bm{\xi}\label{eq:SVMstruct}\\
\text{sb.t.}\; & \min_{\mathbf{y}\in\mathcal{Y}\backslash\mathbf{y}^{k}}\left\{ \mathbf{w}^{\top}\delta\bm{\psi}_{\mathbf{y}}^{k}-\ell_{\mathbf{y}}^{k}\right\} \ge-\xi_{k} & \quad\forall k.\nonumber \end{eqnarray}
In fact, the convex function $o(\mathbf{w})$ can be rewritten as
a sum of a quadratic regulariser and a maximum over an exponentially
sized set of linear functions each corresponding to a particular segmentation
$\mathbf{y}$. 
Which energy functions fit under the umbrella of SVMstruct? In
principle, in the cutting-planes approach \cite{tsochantaridis04SVMstruct}
to solve Eq. \ref{eq:SVMstruct}, we only require efficient and exact
computation of $\arg\min_{\mathbf{y}}E_{\mathbf{w}}(\mathbf{y})$
and $\arg\min_{\mathbf{y}\neq\mathbf{y}^{k}}E_{\mathbf{w}}(\mathbf{y})-\Delta(\mathbf{y},\mathbf{y}^{k})$.
For the scale of images i.e. $n>10^{5}$, submodular energies of the
form $E_{\mathbf{w}}(\mathbf{y})=\mathbf{y}^{\top}\mathbf{F}\mathbf{y}+\mathbf{b}^{\top}\mathbf{y},\: F_{ij}\ge0,b_{i}\in\mathbb{R}$
allow for efficient minimisation by graph cuts. As soon as we include
connectivity constraints as in Eq. \ref{eq:E}, we can only approximately
train the SVMstruct. However some theoretical properties seem to carry
over empirically \cite{finley08intractSVMstruct}.

\subsection{Dynamic SVMstruct with {}``Cheating''}

The SVMstruct does not capture the user interaction part. Therefore,
we add a third term to the objective that measures the amount of \emph{user
interaction} $\iota$ where $\mathbf{u}^{k}\in\{0,1\}^{n}$ is a binary
image indicating whether the user provided the label of the corresponding
pixel or not. One can think of $\mathbf{u}^{k}$ as a partial solution fed
into the system by the user brush strokes. In a sense $\mathbf{u}^{k}$
implements a mechanism for the SVMstruct to \emph{cheat}, because only
the unlabeled pixels have to be segmented by our $\arg\min_{\mathbf{y}}E_{\mathbf{w}}$
procedure, whereas the labeled pixels stay clamped. In the optimisation problem,
we also have to modify the constraints such that the only segmentations
$\mathbf{y}$ \emph{compatible }with the interaction $\mathbf{u}^{k}$
are taken into account. Our modified objective is given by:\begin{eqnarray}
\min_{\bm{\xi}\ge\mathbf{0},\mathbf{w}{\color{darkblue},\mathbf{u}^{k}}} & o(\mathbf{w}{\color{darkblue},\mathbf{U}}):=\frac{1}{2}\left\Vert \mathbf{w}\right\Vert ^{2}+\frac{C}{K}\mathbf{1}^{\top}\bm{\xi}{\color{darkblue}+\iota}\label{eq:SVMstructCheat}\\
\text{sb.t.}\; & \min_{\mathbf{y}\in{\color{darkblue}\mathcal{Y}|_{\mathbf{u}^{k}}}\backslash\mathbf{y}^{k}}\left\{ \mathbf{w}^{\top}\delta\bm{\psi}_{\mathbf{y}}^{k}-\ell_{\mathbf{y}}^{k}\right\} \ge-\xi_{k} & \quad\forall k\nonumber \\
 & {\color{darkblue}\iota\ge\mathbf{a}^{\top}\mathbf{u}^{k}} & {\color{darkblue}\quad\forall k}\nonumber \end{eqnarray}
For simplicity, we choose the amount of user interaction or cheating
$\iota$ to be the maximal $\mathbf{a}$-reweighted number of labeled
pixels $\iota=\max_{k}\sum_{i}a_{i}|u_{i}^{k}|$, with uniform weights
$\mathbf{a}=a\cdot\mathbf{1}$. Other formulations based on the \emph{average}
rather than on the \emph{maximal} amount of interaction proved feasible
but less convenient. We denote the set of all user interactions for
all $K$ images $\mathbf{x}^{k}$ by $\mathbf{U}=[\mathbf{u}^{1},..,\mathbf{u}^{K}]$.
The compatible label set $\mathcal{Y}|_{\mathbf{u}^{k}}=\{0,1\}^{n}$
is given by $ $$\{\hat{\mathbf{y}}\in\mathcal{Y}|u_{i}^{k}=1\Rightarrow\hat{y}_{i}=y_{i}^{k}\}$
where $\mathbf{y}^{k}$ is the ground truth labeling. Note that $o(\mathbf{w},\mathbf{U})$
is convex in $\mathbf{w}$ for all values of $\mathbf{U}$ and efficiently
minimisable by the cutting-planes algorithm. However the dependence
on $\mathbf{u}^{k}$ is horribly difficult -- we basically have to
find the smallest set of brush strokes leading to a correct segmentation.
Geometrically, setting one $u_{i}^{k}=1$ \emph{halves }the number
of possible labellings and therefore removes half of the label constraints.
The optimisation problem (Eq. \ref{eq:SVMstructCheatStrategy}) can
be re-interpreted in two different ways: \\
Firstly, we can define a modified energy $\tilde{E}_{\mathbf{w},\mathbf{v}}(\mathbf{y})=E_{\mathbf{w}}(\mathbf{y})+\sum_{i\in\mathcal{V}}u_{i}^{k}\phi_{i}(y_{i},y_{i}^{k})$
with additional \emph{cheating potentials} $\phi_{i}(y_{i},y_{i}^{k}):=\infty$
for $y_{i}\neq y_{i}^{k}$ and $0$ otherwise allowing to treat the
SVMstruct with cheating as an ordinary SVMstruct with modified energy
function $\tilde{E}_{\mathbf{w},\mathbf{v}}(\mathbf{y})$ and extended
weight vector $\tilde{\mathbf{w}}=[\mathbf{w};\mathbf{u}^{1};..;\mathbf{u}^{K}]$.
\\
A second (but closely related) interpretation starts from the fact
that the true label $\mathbf{y}^{k}$ can be regarded as a \emph{feature
vector} of the image $\mathbf{x}^{k}$%
\footnote{In fact, it is probably the most informative feature one can think
of. The corresponding predictor is given by the identity function.%
}. Therefore, it is feature selection in a very particular
feature space. There is a direct link to multiple kernel learning
-- a special kind of feature selection.

\subsection{Optimisation with strategies}

We explored two approaches to minimise $o(\mathbf{w},\mathbf{U})$.
Based on the discrete derivative $\frac{\partial o}{\partial\mathbf{U}}$,
we tried coordinate descent schemes. Due to the strong coupling of
the variables, only very short steps were possible%
\footnote{In the end, we can only safely flip a single pixel $u_{i}^{k}$ at
a time to guarantee descent.%
}. Conceptually, the \emph{process} of optimisation is decoupled from
the user interaction process, where removal of already known labels
from the cheating does not make sense. At every stage of interaction,
a user acts according to a \emph{strategy} $s:(\mathbf{x}^{k},\mathbf{y}^{k},\mathbf{u}^{k,t},\mathbf{y},\mathbf{w})\mapsto\mathbf{u}^{k,t+1}$.
The notion of strategy or policy is also at the core of a robot
user. In order to capture the sequential nature of the human interaction
and assuming a fixed strategy $s$, we relax Eq. \ref{eq:SVMstructCheat}
to
\begin{eqnarray}
\min_{\bm{\xi}\ge\mathbf{0},\mathbf{w}} & o(\mathbf{w},T):=\frac{1}{2}\left\Vert \mathbf{w}\right\Vert ^{2}+\frac{C}{K}\mathbf{1}^{\top}\bm{\xi}{\color{darkblue}+\iota}\label{eq:SVMstructCheatStrategy}\\
\text{sb.t.}\; & \min_{\mathbf{y}\in{\color{darkblue}\mathcal{Y}|_{\mathbf{u}^{k,T}}}\backslash\mathbf{y}^{k}}\left\{ \mathbf{w}^{\top}\delta\bm{\psi}_{\mathbf{y}}^{k}-\ell_{\mathbf{y}}^{k}\right\} \ge-\xi_{k} & \quad\forall k\nonumber \\
 & {\color{darkblue}\iota\ge\mathbf{a}^{\top}\mathbf{u}^{k,T},\:\mathbf{u}^{k,T}=s^{T}(\mathbf{x}^{k},\mathbf{y}^{k},\mathbf{w})} & {\color{darkblue}\quad\forall k}\nonumber \end{eqnarray}
where we denote repeated application of the strategy $s$ by $s^{T}(\mathbf{x}^{k},\mathbf{y}^{k},\mathbf{w})=\bigcirc_{t=0}^{T-1}s(\mathbf{x}^{k},\mathbf{y}^{k},\mathbf{u}^{k,t},\mathbf{w})$
and by $\bigcirc$ the function concatenation operator. Note that
we still cannot properly optimise Eq. \ref{eq:SVMstructCheatStrategy}.
However, as a proxy, we develop Eq. \ref{eq:SVMstructCheatStrategy}
forward by starting at $t=0$ with $\mathbf{u}^{k,0}$. In every step
$t$, we interleave the optimisation of the convex objective $o(\mathbf{w}^{t},t)$ 
and the inclusion of a new user stroke yielding
$\mathbf{w}^{T}$ as final parameter estimate.

\subsection{Experiments}
\begin{figure}[t]
\begin{centering}
\includegraphics[width=0.95\columnwidth]{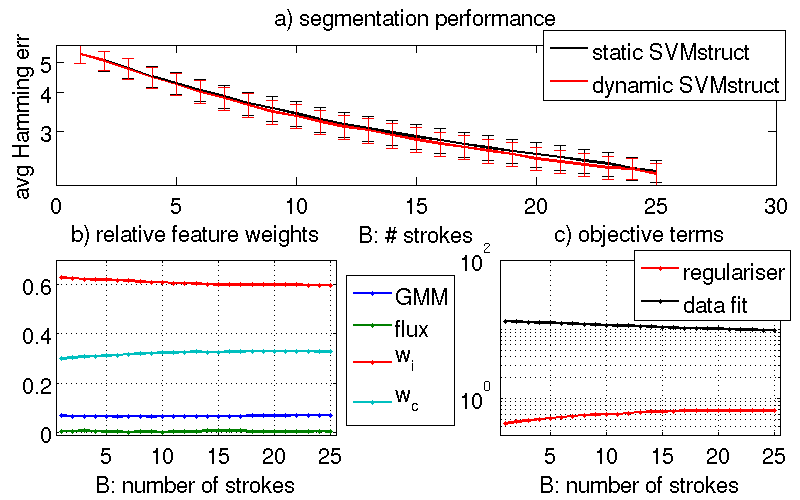}
\par\end{centering}

\caption{\label{fig:maxmargin}{\it Max-margin stat/dyn}: a) Segmentation performance using GCS when parameters are either statically or dynamically learnt. b) Evolution of $\mathbf{w}$ during the optimisation. c) Evolution of the first two terms of $o(\mathbf{w}).$}
\end{figure}

We ran our optimisation algorithm with GCS on 5-fold CV train/test splits of the GrabCut images. We used unary potentials (GMM and flux) as well as two pairwise potentials (Ising and contrast) and the center robot user with $B=25$ strokes. Fig. \ref{fig:maxmargin}b shows, how the relative
weight of the linear parameters varies over time. At the beginning, smoothing (high $w_i$) is needed whereas later, edges are most important (high $w_c$). Also the SVMstruct objective changes. Fig. \ref{fig:maxmargin}c makes clear that the data fit term decreases over time and 
regularisation increases.
However, looking at the test error in Fig. \ref{fig:maxmargin}a (averaged over 5 folds) we see only very little difference between the performance of the initial parameter $\mathbf{w}^0$ and the final parameter $\mathbf{w}^T$.
Our explanation is based on the fact that GCS is too simple as it does not include connectivity or unary iterations.
In addition to the Gaussian Mixture Model (GMM) based color potentials, we also experimented with flux potentials \cite{lepitzky07shapeFitting} as a second unary term. Figure \ref{fig:maxmargin}b shows one example, where we included a flux unary potential. We get almost identical behavior without flux unaries.

\section{Conclusion}
\label{sec:conclusion}

This paper showed how user interaction models (robot users) can be used to train and evaluate interactive systems. We demonstrated the power of this approach on the problem of parameter learning in interactive segmentation systems. We showed how simple grid search can be used to find good parameters for different segmentation systems under an active user interaction model. We also compared the performance of the static and dynamic user interaction models. With more parameters, the approach becomes infeasible, which naturally leads to the max margin framework. 

We introduced an extension to SVMstruct, which allows it to incorporate user interaction models, and showed how to solve the corresponding optimisation problem. However, crucial parts of state-of-the-art segmentation systems include (1) non-linear parameters, (2) higher-order potentials (e.g. enforcing connectivity) and (3) iterative updates of the unary potentials – ingredients that cannot be handled directly inside the max-margin framework. In future work, we will try to tackle these challenges to enable learning of optimal interactive systems. 

{\small
\bibliographystyle{ieee}
\bibliography{iasyslearn}
}

\end{document}